\documentclass[10pt,twocolumn,letterpaper]{article}

\usepackage{iccv}
\usepackage{times}
\usepackage{epsfig}
\usepackage{graphicx}
\usepackage{amsmath}
\usepackage{amssymb}

\RequirePackage{algorithm}
\RequirePackage{algorithmic}
\usepackage[inline,shortlabels]{enumitem,}
\usepackage{multirow}
\usepackage[table]{xcolor}

\usepackage{authblk}
\usepackage{blindtext}

\usepackage[breaklinks=true,bookmarks=false]{hyperref}

\iccvfinalcopy 

\def\iccvPaperID{10895} 

\ificcvfinal\fi

\begin{document}

\title{Building-GAN: Graph-Conditioned Architectural Volumetric Design Generation}

\newcommand\CoAuthorMark{\footnotemark[\arabic{footnote}]}

\author[1]{Kai-Hung Chang\thanks{Contributed equally.}}
\author[1]{Chin-Yi Cheng\protect\CoAuthorMark}
\author[1]{Jieliang Luo}
\author[2]{Shingo Murata}
\author[1]{Mehdi Nourbakhsh}
\author[2]{Yoshito Tsuji}
\affil[1]{Autodesk Research, United States}
\affil[2]{Obayashi AI Design Lab, Japan}

\maketitle
\ificcvfinal\thispagestyle{empty}\fi

\begin{abstract}

Volumetric design is the first and critical step for professional building design, where architects not only depict the rough 3D geometry of the building but also specify the programs to form a 2D layout on each floor. Though 2D layout generation for a single story has been widely studied, there is no developed method for multi-story buildings. This paper focuses on volumetric design generation conditioned on an input program graph. Instead of outputting dense 3D voxels, we propose a new 3D representation named voxel graph that is both compact and expressive for building geometries. Our generator is a cross-modal graph neural network that uses a pointer mechanism to connect the input program graph and the output voxel graph, and the whole pipeline is trained using the adversarial framework. The generated designs are evaluated qualitatively by a user study and quantitatively using three metrics: quality, diversity, and connectivity accuracy. We show that our model generates realistic 3D volumetric designs and outperforms previous methods and baselines.
\end{abstract}

\section{Introduction}
\label{sec:introduction}

\begin{figure}[t]
  \centering
  \includegraphics[width= 0.5\textwidth]{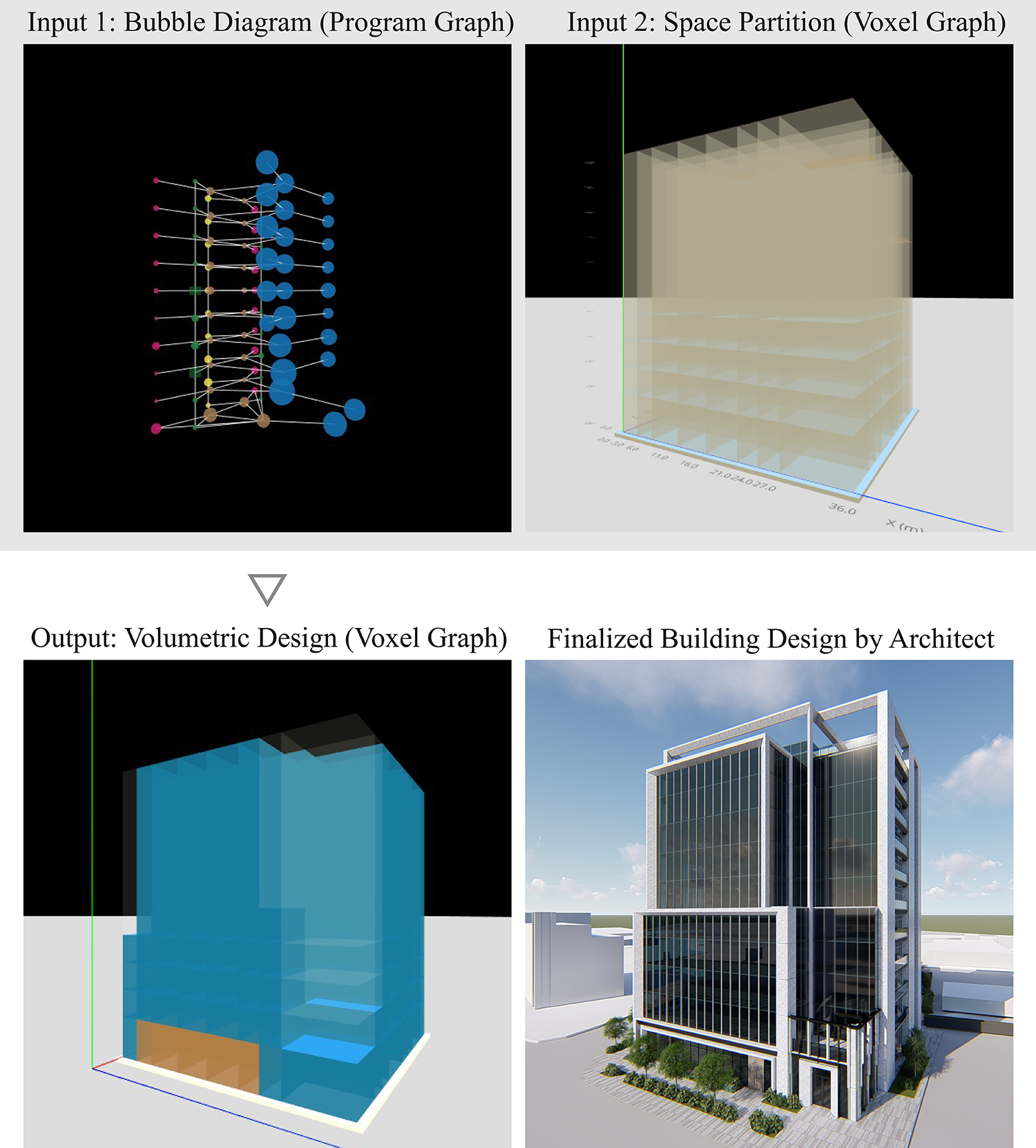}
  \caption{Our model takes in a program graph (also called bubble diagram) and a design space in voxel graph representation, and outputs a variety of volumetric designs. Professional architects can convert the output into detailed building design efficiently.}
  \label{fig:teaser}
\end{figure}

\begin{figure}[t]
  \centering
  \includegraphics[width=0.5\textwidth]{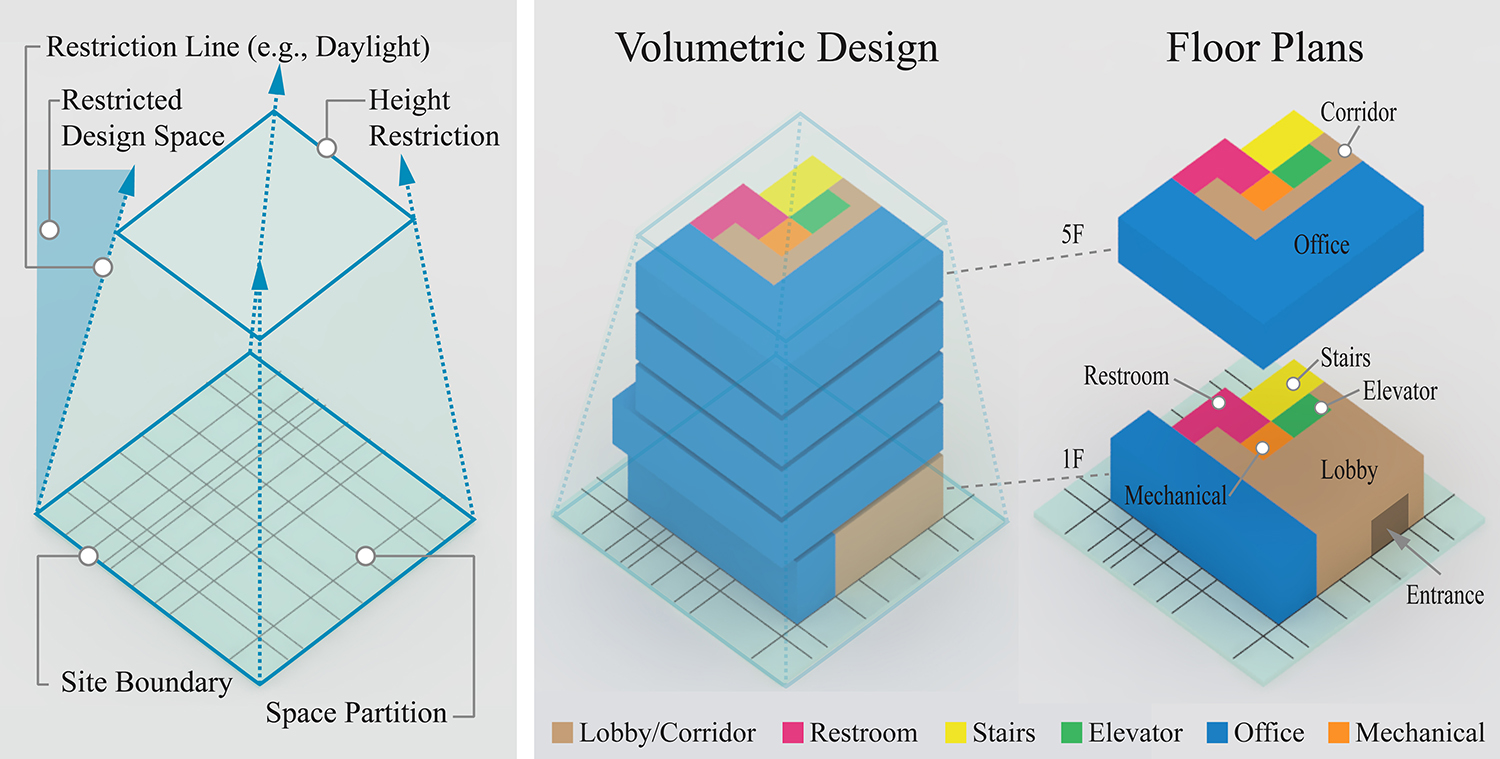}
  \caption{Left: an example of valid design space. Right: an example of volumetric design within the valid design space}
  \label{fig:vol_desgin}
\end{figure}

Volumetric design (also called massing design or schematic design) is the first step when an architect designs a building on a given land site. Based on the local building codes applied to the site, the building can only be designed within a valid design space, which is usually not a regular cuboid. For instance, the daylight restrictions prevent the building from casting too much shadow over its neighboring building by drawing a slant line as upper bound. Within the valid design space, a volumetric design not only depicts the volumetric 3D shape of the building, but also produces 2D program layouts for each story. An example is illustrated in Figure \ref{fig:vol_desgin}. 
The architect then uses the finalized volumetric design to gradually develop all the details for construction, including façade design, interior design, structure systems, etc. While volumetric design is the foundation of the design and construction process, making a good volumetric design usually requires a significant amount of time and effort. An efficient pipeline to generate volumetric design will bring a great impact on the architecture and construction industry.  

Generating realistic 2D room layouts has been a popular topic for many years. Existing methods include optimization-based  \cite{merrell2010computer, bymw_goodLayout_sigg13} and learning-based \cite{wu2019data, nauata2020house, hu2020graph2plan, di2020end} approaches. Recently, researchers start looking at how to integrate program graphs into layout generation tasks using graph neural networks (GNNs) \cite{nauata2020house, hu2020graph2plan, di2020end}. Program graph, also called bubble diagram, is a graph that illustrates the relations between programs or rooms and is a common representation used by professional architects to explore design ideas. Similar to House-GAN \cite{nauata2020house}, this paper also focuses on the graph-conditioned layout generation task. The task requires the output layouts to be compatible to the condition input \textit{program graphs}. However, there is no literature on extending the task to 3D. Our goal is to produce multiple layouts, which stack up and form a volumetric design for a multi-story building.

Though it might seem straight-forward to transfer previous 2D approaches to 3D, there are several challenges and limitations when applying previous approaches: 
\begin{itemize}
    \item Compared to the 2D counterparts, 3D program graphs are not only larger in size, but also more complex with additional inter-story relations. The output design space also increases by the number of stories. 
    \item The raw rasterized output used in previous works cannot produce clean corners and edges due to the fine discretization of pixels. For instance, boundaries are usually jagged, rooms can be poorly aligned and overlapping each other, there might be small dents or bulges in some rooms, etc.
    \item Volumetric images (usually defined as 3D regular grids with uniformly discretized voxels) have the closest structural similarity to rectangular buildings than other 3D representations, such as point clouds or meshes. 
    However, it is not computational and memory efficient to use this dense representation for polygonal rooms.  
    Moreover, there are voxels within the regular grid but not in the irregular valid design space that take unneeded memory and computation. 
    
\end{itemize}

%

To overcome these challenges and limitations, we 
propose \textit{voxel graph}, a novel 3D representation that can encode irregular voxel grids with non-uniform space partitioning. To bridge between the input \textit{program graph} and the output \textit{voxel graph}, we design a pointer-based cross-modal modules in our generative adversarial graph network. The pointer module can be used not only for message passing, but also as a decoder to output probability over a dynamic set of valid programs. 

We also work with professional architects to create a synthetic dataset that contains 120,000 volumetric designs based on realistic building requirements. We evaluate our model qualitatively and quantitatively, and it outperforms existing method by a large margin in all the three metrics: quality, diversity, and connectivity accuracy. 

In summary, our main contributions are: 1) a new 3D representation, \textit{voxel graph}; 2) a graph-conditioned generative adversarial network (GAN) using GNN and pointer-based cross-modal module; 3) an automated pipeline to generate valid volumetric designs through simple interaction; and 4) a synthetic dataset that contains 120,000 volumetric design and their corresponding program graphs. We will share the code, model, and dataset. 

\section{Related Work} 
\label{sec:related work}
\subsection{Voxel Representations}
Regular grid representation using voxels, such as occupancy grids, has been studied since the 3D extension of 2D convolution. To achieve 3D shape synthesis, researchers build encoder-decoder models, such as deep belief network \cite{wu20153d}, variational auto-encoder (VAE) \cite{kingma2013auto}, generative adversarial network (GAN) \cite{wu2016learning, smith2017improved}, and energy-based model \cite{xie2018learning}. However, due to the dense representation for sparse occupancy, voxel representation is notorious for its cubic computational cost and poor scalability to higher resolutions and larger sizes. Existing methods to mitigate the problem include sparse convolution \cite{graham2017submanifold, graham20183d, choy20194d} and octree representation \cite{riegler2017octnet, wang2017cnn, wang2018adaptive}. 

Our proposed voxel graph combines voxel-based and graph-based representations by encoding voxels into graph nodes. Similar idea was proposed in Point-Voxel CNN \cite{liu2019point}. To enhance the local modeling capability, it has a high-resolution point-based branch as well as a low-resolution voxel-based branch for point cloud encoding. Another feature of our voxel graph is the ability to support non-uniform space partition. Polyfit \cite{nan2017polyfit, fang2020connect} reconstructs 3D models by selecting space partition planes extracted from point clouds. BSP-Net \cite{chen2020bsp} learns to generate compact meshes using binary space partitioning. NeuralSim and NeuralSizer \cite{pmlr-v119-chang20a} also use graphs to represent structure grids (i.e., columns and beams) of buildings instead of dense voxels.





\begin{figure*}[h]
  \centering
  \includegraphics[width=.9\textwidth]{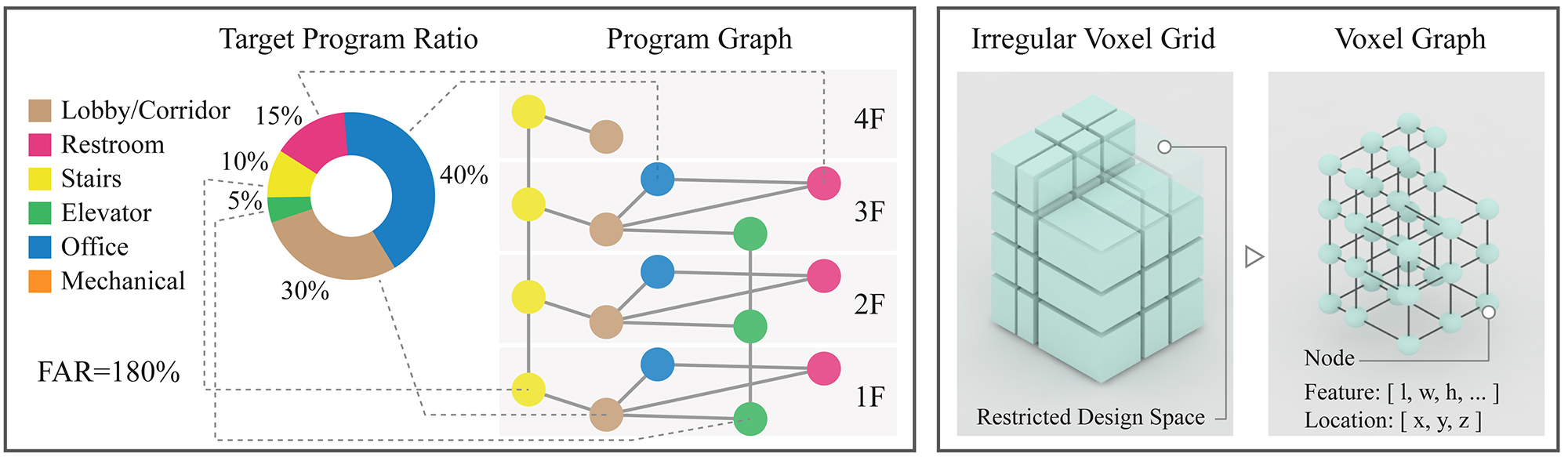}
  \caption{Left: the hierarchical program graph. Right: the irregular grid with non-uniform voxel size and the equivalent voxel graph. }
  \label{fig:graph_representation}
\end{figure*}

\subsection{Graph-conditioned Layout Generation}

To the best of our knowledge, there is no prior works on learning-based 3D layout generation. Alternatively, we review several work on graph-conditioned 2D layout generation. Graph2Plan \cite{hu2020graph2plan} generates bounding boxes for each room, and refines box locations with a cascaded refinement network. The input graphs are retrieved based on user constraints and outline similarity. The user can get various layouts by feeding different graphs, but the model itself cannot produce variation. House-GAN \cite{nauata2020house} proposes a graph-conditioned GAN, where the generator and discriminator are built upon relational architecture - ConvMPN \cite{zhang2020conv}. Xinhan Di \etal \cite{di2020end} uses a similar adversarial approach on interior design with doors, windows, and furniture. LayoutGMN \cite{patil2020layoutgmn} learns to predict structural similarity between two layouts with an attention-based graph matching network. Wamiq Para \etal \cite{para2020generative} explores the idea of generative modeling using constraint generation for layouts. 


\section{Representation and Data Collection}
The goal of this paper is to generate 3D volumetric designs given a program graph and a valid design space. The program graph illustrates the intra-story and inter-story relations between programs. Besides program graph and valid design space, there are other design conditions that are considered by architects in industry practice. Floor area ratio (FAR, derived by dividing the total area of the building by the total area of the parcel), should not exceed a regulation limit. In addition, target program ratio (TPR) defines the approximate ratio between programs. For example, office : corridor : restroom : elevator : stairs $= 50:20:15:5:10$. Both TPR and FAR are encoded into the program graph as described in Section \ref{sec: hpg} and are used as the model input.


Another input is a valid design space, which may be irregular due to building codes. The design space can be further partitioned freely based on architect's decisions or statistical heuristics. In practice, before starting the design process, architects usually partition the space by considering construction standards, structure systems, and conventional modules. Inspired by this partitioning process, we invent the representation, \textit{voxel graph}, as described in Section \ref{sec: vg}.

\subsection{Data Collection}
Since there is no publicly available dataset for volumetric designs from real buildings, we create a synthetic dataset with 120,000 volumetric designs for commercial buildings using parametric models. The site of each design is bounded within $40\times40\times50  m^3$, where different site conditions are randomly generated. The heuristics behind the parametric models are based on the rules and knowledge provided by professional architects. Although these parametric models are able to explore possible volumetric designs, they are not capable of fitting the constraints. Therefore, we generate the designs first and then compute the voxel graph, program graph, FAR, and TPR for each design. Please refer to the supplementary for more details and visualization of the synthetic dataset. The dataset can also be used to explore other learning-based design tools or relevant tasks in computer vision and graphics.

\subsection{Hierarchical Program Graph}
\label{sec: hpg}
Given a building datum, we first construct 2D program graphs for each story. Each program node feature includes the program type and the story level. Here, we consider 6 program types: lobby/corridor, restroom, stairs, elevator, office, and mechanical room. A program edge shows the two programs are connected by a door or opening. To construct the 3D program graph, we stack all 2D program graphs and chain the stairs and elevators, since they are the only paths for moving vertically. In practice, the 3D program graph also represents the circulation of the building. 

Recall that there are two other design condition inputs: FAR and TPR. The FAR limit is stored as a graph-level feature. As for TPR, we add one hierarchy on top of the 3D program graph. We create one master program node for each program type and connect them to all program nodes of the same type. The edges allow the master node to allocate different area sizes on each program node through message passing. Please refer to left of Figure \ref{fig:graph_representation}.

\begin{figure*}[h]
  \centering
  \includegraphics[width=\textwidth]{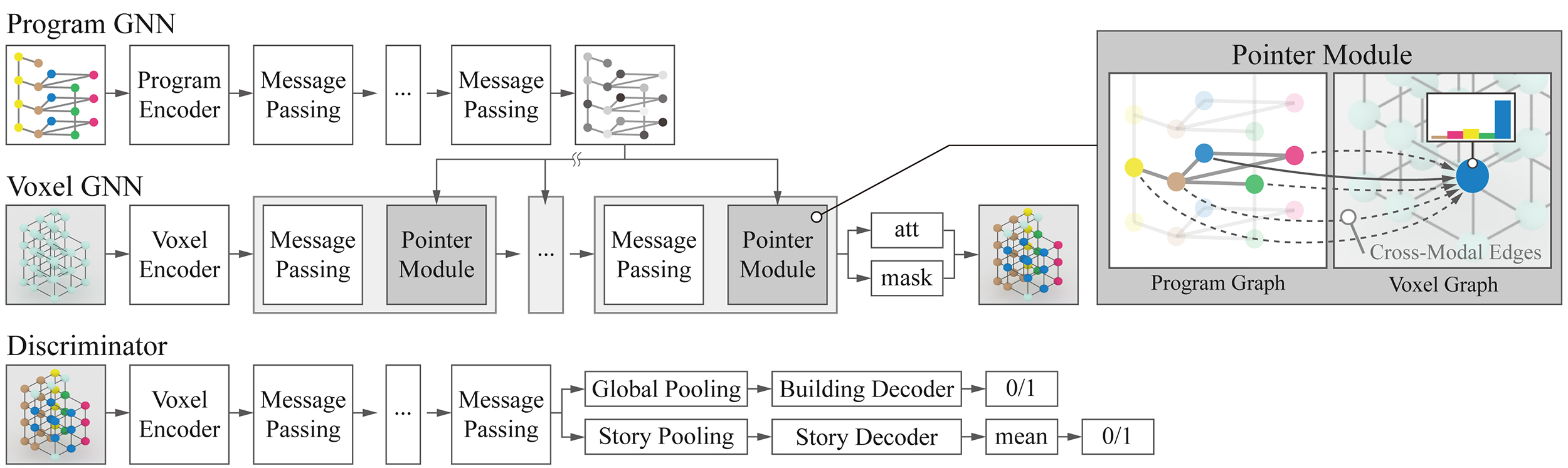}
  \caption{An overview of Building-GAN. Top: the Program GNN, Voxel GNN, and Cross-Modal Pointer Module for the generator. Bottom: the discriminator with the building and story level decoders.}
  \label{fig:network_structure}
\end{figure*}

\subsection{Voxel Graph}
\label{sec: vg}


To overcome the challenges and limitations listed in Section 
\ref{sec:introduction}, we invent a 3D representation called \textit{voxel graph}. Each node represents a voxel and the voxel information (coordinate and dimension) is stored as node features. Different from volumetric images with voxel grids, voxel graph does not assume regular grids and consumes memory only for occupied voxels. Moreover, it allows non-uniform space partitioning, which avoids over-discretization when using the uniform voxel size.

Theoretically, voxel nodes can encode arbitrary 3D primitives, but in this paper, only cuboids with varying sizes are used to build up the approximated valid design space. When parsing the data, the space partition is defined by the projection of all 2D layouts. In real-world practice, walls tend to align across different stories for structural stability or construction considerations, which leads to a reduced amount of voxels in the space partition. Next, we turn the voxels into graph nodes and store the voxel information (location and dimension) as node features and program type as node labels. Node mask is also stored in case of nodes that are left unused and does not have any program type. Lastly, a voxel edge connects two voxel nodes if they share a face. The final voxel graph should look like an irregular cubic lattice as illustrated in the right of Figure \ref{fig:graph_representation}.


\section{Method}
\label{sec: method}


We formulate the framework as a graph-conditioned GAN. The generator is composed by two GNNs for the program graph and voxel graph, connected by a cross-modal pointer module. The discriminator is composed by a GNN with two decoders to evaluate design from both building and story level. An overview of our model is illustrated in Figure \ref{fig:network_structure}.


\subsection{Generator}
\label{sec: gen}
\subsubsection{Program GNN}
Our generator starts with a program graph neural network to encode the input program graph. Denote random program noise as $z^{p}$, FAR limit as $F$, program node feature $i$ as $x_{i}$, neighbor of node $i$ as $Ne(i)$, node cluster of $i$'s program type as $Cl(i)$, target program ratio of $i$'s program type as $r_{Cl(i)}$, multi-layer perceptron as $MLP$, mean pooling as $Mean$, and concatenation operator as $[\cdot,\cdot]$. We first map the node feature to the embedding space (\ref{eq:1}), then compute message passing $T$ times. In each message passing step, we compute the message from neighboring nodes (\ref{eq:2}) and mean pool all nodes with the same program type as the master node embedding (\ref{eq:3}). Lastly, we update the node embeddings with residual learning to avoid gradient vanishing (\ref{eq:4}). After $T=5$ steps of message passing, the final embedding of program node $i$ is denoted as $x^{T}_{i}$.

\begin{equation}
    x_{i}^{0} = MLP_{enc}^{p}([x_{i}, z^{p}_{i}, F])
    \label{eq:1}
\end{equation}
\begin{equation}
    m_{i}^{t} = \frac{1}{|Ne(i)|}\sum_{j\in Ne(i)} MLP_{message}^{p}([x_{i}^{t}, x_{j}^{t}]) 
    \label{eq:2}
\end{equation}
\begin{equation}
    c_{i}^{t} = Mean_{j\in Cl(i)} (\{x_{j}^{t}\}) 
    \label{eq:3}
\end{equation}
\begin{equation}
    x_{i}^{t+1} = x_{i}^{t} + MLP_{update}^{p}([x_{i}^{t}, m_{i}^{t}, r_{Cl(i)}c_{i}^{t}, F])
    \label{eq:4}
\end{equation}

\subsubsection{Voxel GNN}
The input voxel features $v_{k}$ and voxel noise $z^{v}_{k}$ are first encoded by the voxel GNN encoder. To better encode the story index, we choose positional encoding (PE) as proposed in \cite{vaswani2017attention} and add it to the processed embedding (\ref{eq:5}). Instead of appending the absolute coordinates in voxel features, we use the relative displacements $p_{k}-p_{l}$ in message computation (\ref{eq:6}). Voxel node embeddings are updated with residual learning (\ref{eq:7}). 

\begin{equation}
    v_{k}^{0} = MLP_{enc}^{v}([v_{k}, z^{v}_{k}]) + PE(story_{k})
    \label{eq:5}
\end{equation}
\begin{equation}
    n_{k}^{t} = \sum_{l\in Ne(k)} MLP_{message}^{v}([v_{k}^{t}, v_{l}^{t}, p_{k}-p{l}]) 
    \label{eq:6}
\end{equation}
\begin{equation}
    v_{k}^{t} = v_{k}^{t} + MLP_{update}^{v}(v_{k}^{t}, n_{k}^{t})
    \label{eq:7}
\end{equation}

\subsubsection{Pointer-based Cross-Modal Module}
After processing the program graph with the program GNN, the final embedding of program nodes can be viewed as the virtual "blueprint" of a design. Therefore, it is necessary to "look" at this blueprint to generate the output. To bridge between the program graph and the voxel graph, we introduce a pointer-based cross-modal module. Inspired by the application \cite{see2017get,nash2020polygen} of the Pointer Network \cite{vinyals2015pointer} in natural language processing and mesh generation tasks, we construct a pointer module to achieve message passing between the voxel nodes and all the program nodes on the same story. We cannot use a fixed length output to model program type distribution since 1) different stories can have different numbers of program nodes to choose from, for example, one floor has five rooms and another one has seven rooms; and 2) if there are two program nodes with the same program type, we want to differentiate between the two nodes, such as two restrooms in the same floor.

The pointer module returns three terms: $mask_{k}$, $att_{k}$, and $v_{k}^{t+1}$ (\ref{eq:8}). $mask_{k}$ is used as a soft prediction whether the voxel node $k$ is used or not (\ref{eq:9}). If it is not used, it is left unused and has no program type. Otherwise, $att_{k}$ is the attention distribution over the set of program nodes on the same floor (\ref{eq:10}, \ref{eq:11}).  An updated embedding $v_{k}^{t+1}$ is computed by the weighted sum of the program embeddings $x^{T}_{i}$ multiplied by the soft prediction $mask_{k}$ with residual learning (\ref{eq:12}). 
\begin{equation}
    mask_{k}, att_{k}, v_{k}^{t+1} = Pointer(v^{t}_{k}, \{x^{T}_{i}\})
    \label{eq:8}
\end{equation}
\begin{equation}
    mask_{k} = \sigma(MLP(v^{t}_{k}))
    \label{eq:9}
\end{equation}
\begin{equation}
    e_{k,i} = \theta ^{T} \tanh(W_{x}x^{T}_{i} + W_{v}v^{t}_{k})
    \label{eq:10}
\end{equation}
\begin{equation}
    att_{k} = \textrm{gumbel softmax}(e_{k}) 
    \label{eq:11}
\end{equation}
\begin{equation}
    v^{t+1}_{k} = v^{t}_{k}+mask_{k} \sum_{i}att_{k, i}x^{T}_{i}
    \label{eq:12}
\end{equation}

We experiment different ways to integrate the pointer module. It can be placed after every several message passing steps in voxel GNN. Our baseline model uses 12 steps of message passing and call the pointer module once every 2 steps. Please refer to the supplementary for the complete model and algorithm. Conceptually, these pointer modules should gradually improve the design. Note that the output $att_{k}$ indicates which program node is associated to the program type of the voxel node, instead of merely the program type prediction.

\subsection{Discriminator}
\label{sec: dis}
Our discriminator is trained to distinguish if a given design is generated by the generator or sampled from the dataset. Therefore, we take a similar architecture as voxel GNN, but without using the pointer modules. The program type predictions are concatenated to the encoded voxel node features. After $T=12$ message passing steps, two separate decoders are used. A graph-level max-pooling decoder evaluates the design as a whole while a story-level max-pooling decoder evaluates the per-story layouts individually. 

\begin{equation}
    o^{global} = MLP^{dec}_{global}(\sum_{k} v^{T}_{k})
    \label{eq:13}
\end{equation}
\begin{equation}
    o^{story} = Mean_{story \; s}(MLP^{dec}_{story}(\sum_{k \in s} v^{T}_{k}))
    \label{eq:14}
\end{equation}

\subsection{Loss}

We use the WGAN-GP \cite{gulrajani2017improved} loss with gradient penalty set to 10. The two decoder outputs from the discriminator are equally weighted. The gradient penalty is computed by linearly interpolating the cross-modal attention between real data and generated output, while fixing the voxel graph connectivity. 


\subsection{Evaluation Metric}
We evaluate the generated design in terms of quality,  diversity, and connectivity accuracy. The quality and diversity of the output design is evaluated with the Fr\'echet Inception Distance (FID) score \cite{heusel2017gans}. FID score has demonstrated high correlation to human judgement and has been widely used in many 2D and 3D studies. Our reference model is based on a larger version of 3D Descriptor Net \cite{xie2018learning}. We replace all convolution layers with 6 residual blocks due to the higher complexity of our data. Then we flatten the embedding to a 128-dimension tensor using convolution operation and pass it to a dense layer for loss computation. The FID score is measured over 10,000 samples. We also run a user study with architects to measure the quality in Section \ref{sec:user_study}.

The connectivity accuracy (Con.) is measured by the number of the program (room) connections observed from both the generated design and in the program graph, divided by the amount of all edges in the program graph. Note that only when two rooms are connected in the program graph but disconnected in the voxel graph, it is considered as inaccurate, since there is no shared wall to put a door. It is accurate when two rooms are connected in voxel graph but disconnected in program graph, because designers can decide not to put a door on the shared walls. 

For more details about model implementation, hyper parameters, training environment, and user study, please refer to the supplementary.
\section{Experiments}

\begin{figure*}[h]
  \centering
  \includegraphics[width=0.7\textwidth]{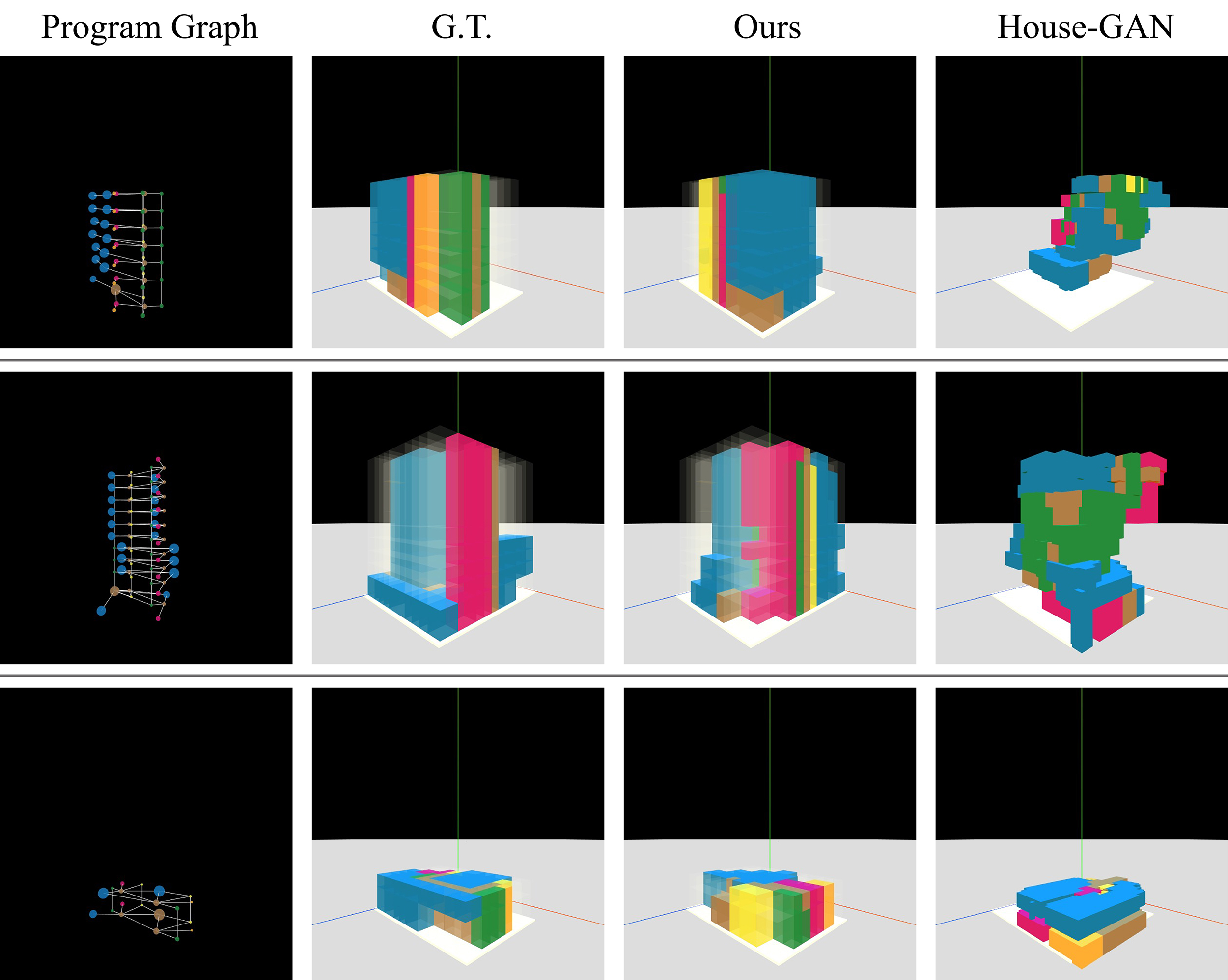}  
  \caption{For each program graph, volumetric designs are generated by our model and by House-GAN \cite{nauata2020house}.}
  \label{fig:viz_result}
\end{figure*}

\subsection{Baseline Comparison and Visualization}
We compare our model to a slightly modified version of House-GAN \cite{nauata2020house}.  A major difference between our model and House-GAN is that House-GAN directly generates layout masks of size $40 \times 40$ on the nodes of the program graph. Since House-GAN does not use voxel graph representation, it assumes that the valid design space is a regular grid. House-GAN discriminator places the generated masks back to the program nodes as features and determines if the program graph is valid. To extend the House-GAN to 3D, we append the story index of each program node to its feature. 

Figure \ref{fig:viz_result} shows example designs from ground truth, our model, and House-GAN. Our model shows capability of generating designs that have realistic 3D shapes and clean facade surfaces.  We also observe that the number of jagged boundaries are reduced due to the usage of voxel graph representation. In addition, the layouts of individual stories are reasonable. For instance, the functional programs such as elevators, stairs, and restrooms are arranged as clusters and connected by corridors. Last but not least, the functional programs are nicely aligned in the vertical direction. In contrast, though House-GAN seems to generate reasonable layouts story-by-story, they don't align well when stacked vertically. Quantitative results are presented in Table \ref{table:baseline_comparison}. Our model outperforms House-GAN in both FID and connectivity accuracy. 

\begin{table}[h]
\begin{center}
\begin{tabular}{|l|c|c|c|}
\hline
Method & Parameter & FID & Con.\\
\hline\hline
House-GAN & - & 17.6003 & 0.403\\
\hline
Ours & - & \cellcolor{lightgray} 0.0845  & \cellcolor{lightgray}0.569\\
\hline \hline
\multirow{5}{*}{
\begin{tabular}{@{}l@{}} Voxel Layer \\ (Pointer Frequency \\ = every 2 steps)\end{tabular}
} & 4 & 1.0463 & 0.432\\
& 6 & 0.4139 & 0.501\\
& 8 & 0.2365 & 0.497\\
& 10 & 0.0997 & 0.534\\
& 12 & \cellcolor{lightgray} 0.0845  & \cellcolor{lightgray} 0.569\\ \hline  
\multirow{4}{*}{
\begin{tabular}{@{}l@{}}Pointer Frequency \\ (Voxel Layer = 12)\end{tabular}
} & first + last & 3.3818 & 0.578\\ 
& every 6 steps& 0.2179 & 0.547\\ 
& every 3 steps & 0.1473 & 0.541\\
& every 2 steps & \cellcolor{lightgray} 0.0845 & \cellcolor{lightgray} 0.569\\  
\hline
\end{tabular}
\end{center}
\caption{Quantitative evaluation using FID score and connectivity accuracy. We compare our baseline model to House-GAN. We also experiment baseline models with different numbers of voxel layer and pointer frequencies.}
\label{table:baseline_comparison}
\end{table}


\begin{figure*}[h]
  \centering
  \includegraphics[width=\textwidth]{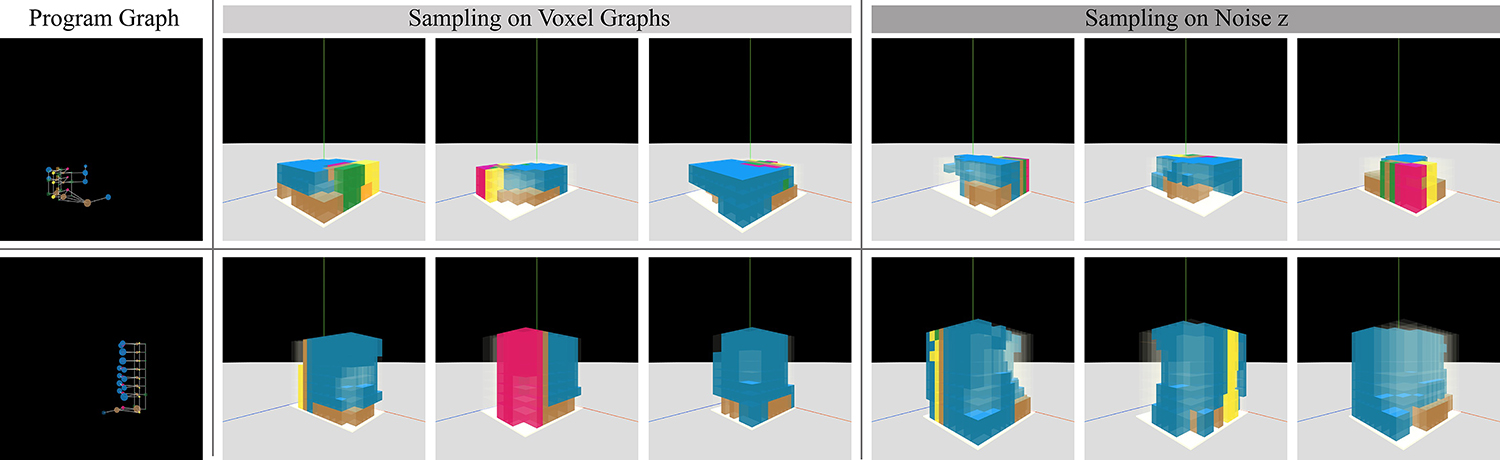}  
  \caption{Design variations generated by fixing the program graph while changing the voxel graph and noise.}
  \label{fig:variation}
\end{figure*}

In Table \ref{table:baseline_comparison}, we also compare models with different hyper-parameter set-ups. First, we fix the frequency of applying the pointer module to 2 and experiment different numbers of message passing layers in voxel GNN. The result shows that using 12 voxel layers yields best performance in both FID and connectivity accuracy. This is not surprising: using more voxel layers allows computing longer-range relations between voxel nodes, which is especially critical for achieving vertical alignment in taller buildings. Next, we fix the number of voxel layer and evaluate the impact of different pointer frequency. The model (first + last) which uses the pointer module only before and after message passing fails to converge. We also run the pointer module every 2, 3, 6 message passing steps in voxel GNN. Using the pointer module every 2 message passing steps yields the best performance in both FID and connectivity accuracy. Reviewing the program graph multiple times during the message passing process might ensure that the information from the program graph is always considered and provide shorter paths for gradient backpropagation.

\subsection{Variation Study}

In Figure \ref{fig:variation}, we visualize examples generated by fixing the program graph while changing the space partition in the voxel graph and noise. The model is able to generate different designs with different patterns, orientations, etc. based on the given noise and design space partition. 

\subsection{Ablation Study}
We run ablation studies on discriminator, positional encoding, and relative position. The results are summarized in Table \ref{table:ablation}. We found that it is necessary to use both story discriminator and building discriminator as using only one leads to inferior performance. Though having message-passing in the voxel GNN, story discriminator has difficulty evaluating inter-story relations and the overall 3D geometry. The better connectivity accuracy with * actually results from noisy low-quality outputs where the layouts are fragmented. Building discriminator proves to play a more crucial role in learning the task, but adding story discriminator significantly improves the output design quality. We also show that using positional encoding (PE) to encode story indices performs better than directly using it (i.e. $1,2\dots$). Lastly, training without relative position (RP) ends up generating low quality designs. It shows that relative position is an indispensable component for our model, since it can capture the direct spatial relationships between connected voxels. 

\begin{table}[h]
\begin{center}
\begin{tabular}{|l|c|c|}
\hline
Ablation Study & FID & Con.\\
\hline\hline
Ours & 0.0845 & 0.569\\ 
\hline\hline
Story discriminator only & 6.8061 & *0.777 \\
Building discriminator only & 1.0464 & 0.459\\
\hline\hline
No PE & 0.1512 & 0.507\\ 
No PE + No RP &  0.8333 & 0.489\\ 
\hline
\end{tabular}
\end{center}
\caption{Ablation study results on discriminator, positional encoding (PE), and relative position (RP). * The higher accuracy here is caused by fragmented low-quality outputs. }
\label{table:ablation}
\end{table}


\subsection{Intermediate Results}
\begin{figure*}[t!]
  \centering
  \includegraphics[width=\textwidth]{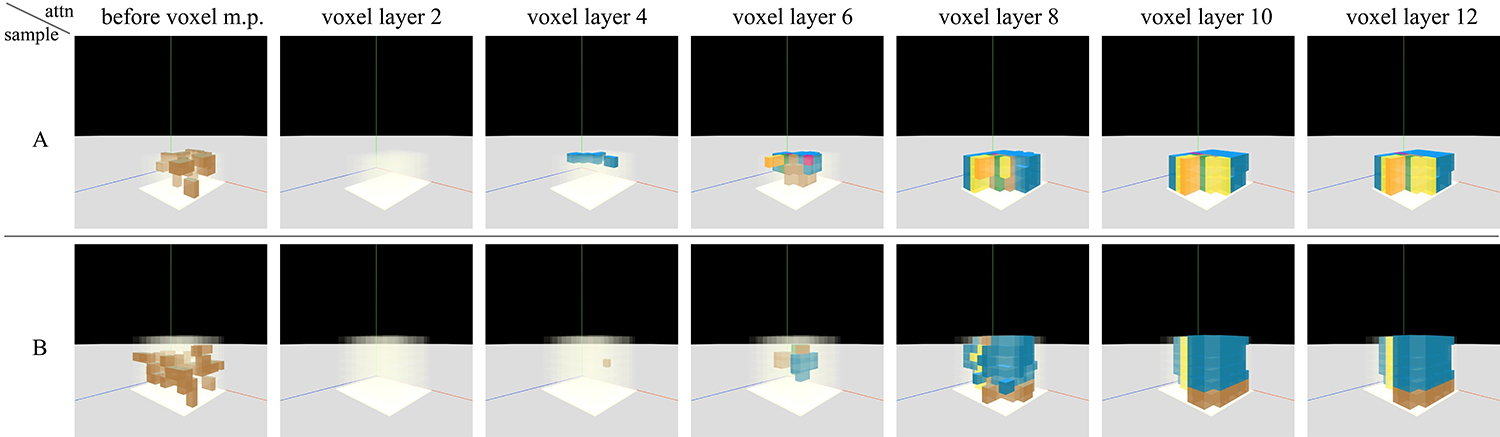}  
  \caption{Visualization of designs generated by all pointer modules for every 2 message passing steps in Voxel GNN during inference.}
  \label{fig:intermediate}
\end{figure*}

In voxel GNN, we run 12 message passing layers and use the pointer module every 2 layers. Since every mask and attention computed by the pointer modules represents a design solution, we are curious to see the "design process" of our model by visualizing the intermediate designs during inference. As shown in Figure \ref{fig:intermediate}, before the voxel message passing, the first attention initializes a seemly random design, trying to allocate only the lobby/corridor type. It makes sense since the decision is only based on the program graph and individual voxel nodes. Interestingly, starting from the second attention, the model chooses to start over and gradually grow the voxels. This behavior aligns with the message passing process since the information from a far distance will flow in with more passing steps. The model also tries to refine the design by overwriting some of the decisions made in previous steps. For example, in the first row of Figure \ref{fig:intermediate}, at layer 6, the isolated restroom (in magenta color) is eliminated at layer 8. 

\subsection{User Study}
\label{sec:user_study}
\begin{figure}[t]
  \centering
  \includegraphics[width= 0.2\textwidth]{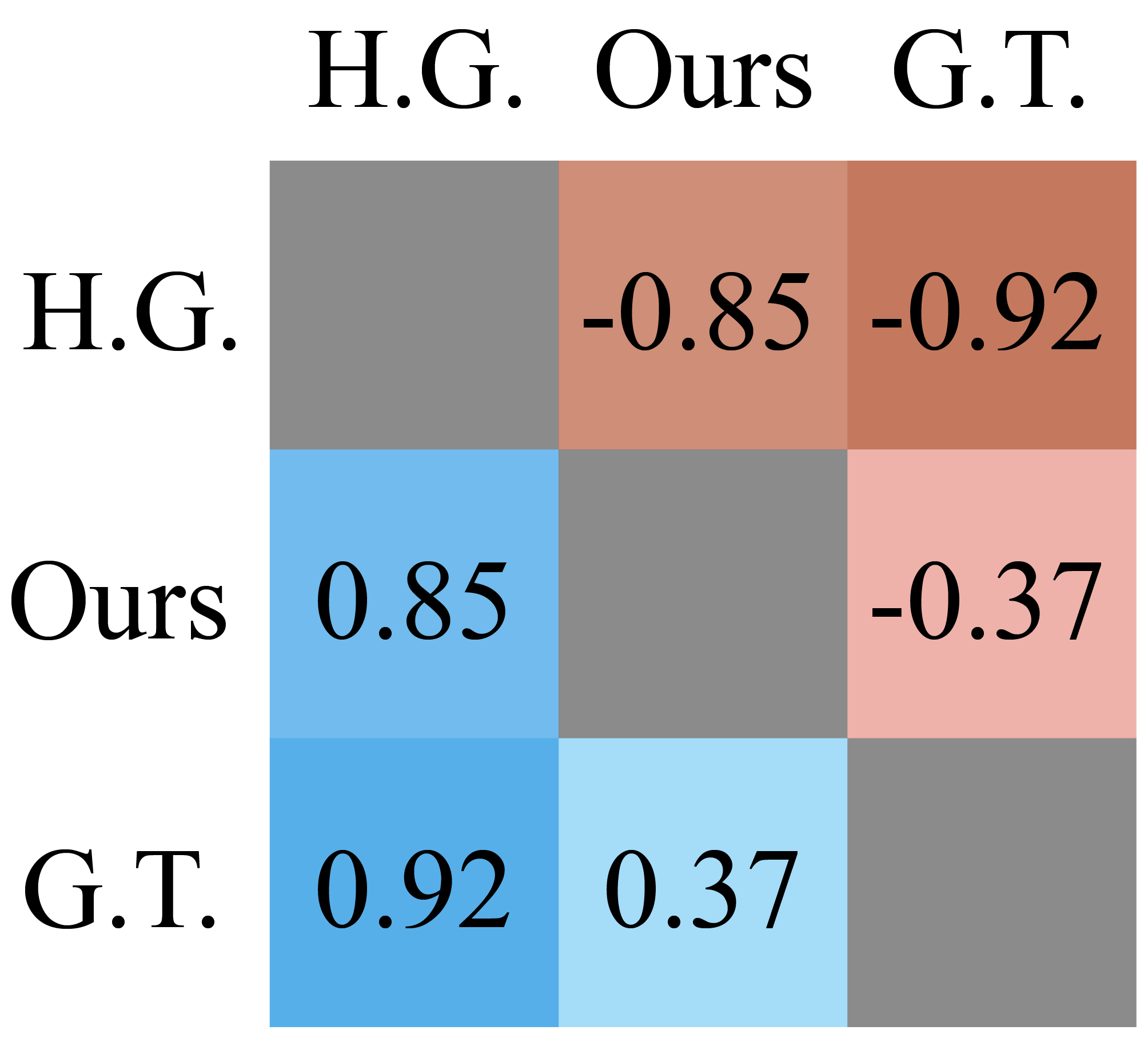}
  \caption{The pairwise quality scores between ground truth (G.T.), our model, and House-GAN(H.G.).}
  \label{fig:user_study}
\end{figure}

To further examine the quality, we conduct user study with 20 professional architects. Each architect is given 48 design pairs that cover all the combinations of the ground truth, House-GAN, and our model. Given a pair of designs, the better design gets 1 point while the worse one gets -1 point. If it's a tie, both get 0 scores. The average  score of a method should range between 1 and -1. The results are shown in Figure \ref{fig:user_study} and it should be read row-by-row. Our model and ground truth defeats House-GAN with scores 0.85 and 0.92 respectively. The ground truth score is only 0.37 when compared to our model, which means in many cases architects cannot clearly tell the difference between the ground truth and ours. 

\subsection{Case Study}
To understand if our pipeline can be useful to the professional building design process, we invite an architect to create the volumetric design using our pipeline and then complete a detailed building design. As shown in the Figure~\ref{fig:teaser}, the results are realistic and aesthetically appealing. The user do feel the pipeline largely increased the efficiency of the design process. For the creation process and detailed feedback, please refer to the supplementary. 


\subsection{Failure Case}

\begin{figure}[t]
  \centering
  \includegraphics[width= 0.4\textwidth]{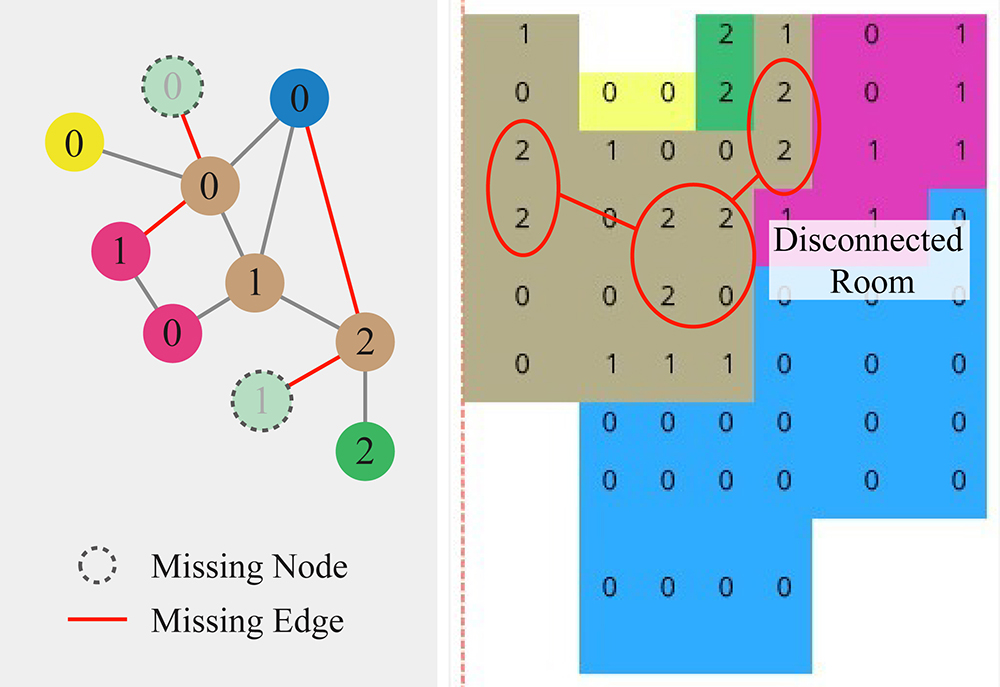}
  \caption{Visualation of the three common flaws in generated volumetric designs. Left: the input program graph. Right: a floor plan in the generated volumetric design given this program graph}
  \label{fig:failure}
\end{figure}


We observe three types of common flaws in the volumetric designs generated from our model: 1) missing nodes; 2) missing edges; and 3) disconnected rooms, as visualized in Figure \ref{fig:failure}. One potential cause of these flaws is that our discriminator only considers the program type on each voxel node instead of the attention between voxel nodes and program graph. Therefore, the discriminator lacks information regarding the specific program nodes which the voxel nodes point to. 
Some of our failed attempts to resolve these flaws are introduced in the supplementary material and we leave the solution for future work.

\section{Conclusion}

In this paper, we try to provide a novel pipeline, Building-GAN, to improve the efficiency on a realistic professional task, volumetric design in the architectural and construction industry. We invent a 3D representation, voxel graph, to represent building designs, and design a generator with a cross-modal pointer module to connect the program graph and voxel graph. Our extensive evaluations, including user testing and user study, show that architects can create numerous valid and valuable designs by interacting with Building-GAN. Future works include enforcing the constraints, such as connectivity, TPR, and FAR, as well as extending the voxel graph for non-cuboid geometries. We will release our code, model, and dataset, and invite the research community to work together on design-related problems in the industries.

{\small
\bibliographystyle{ieee_fullname}
\bibliography{main}
}

\clearpage
\section*{Appendix}

\section{Dataset}
\begin{figure}[h]
  \centering
  \includegraphics[width=0.43\textwidth]{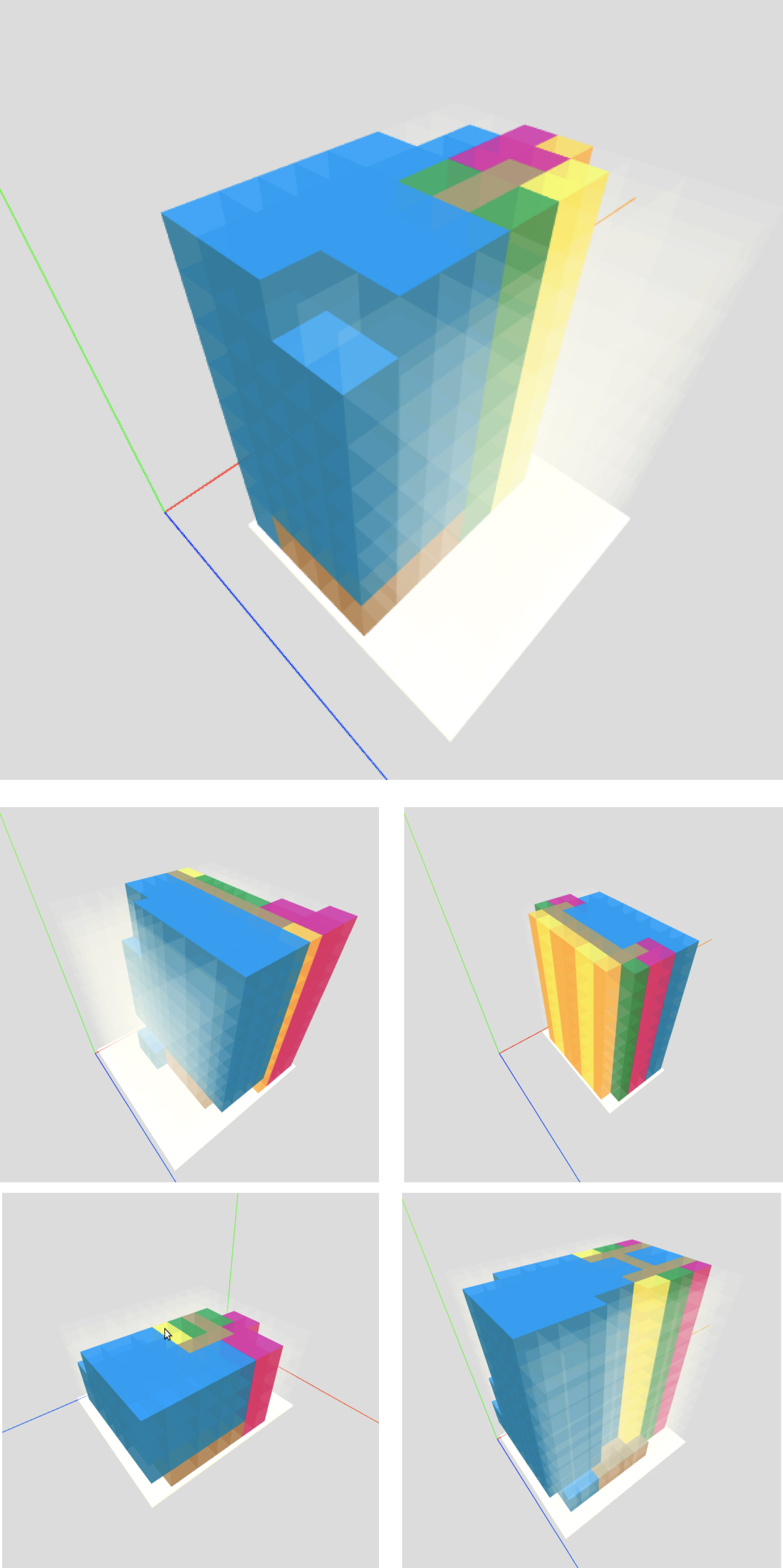}
  \caption{Volumetric design samples in the synthetic dataset}
  \label{fig:syn_1}
\end{figure}

\begin{figure}[h]
  \centering
  \includegraphics[width=0.42\textwidth]{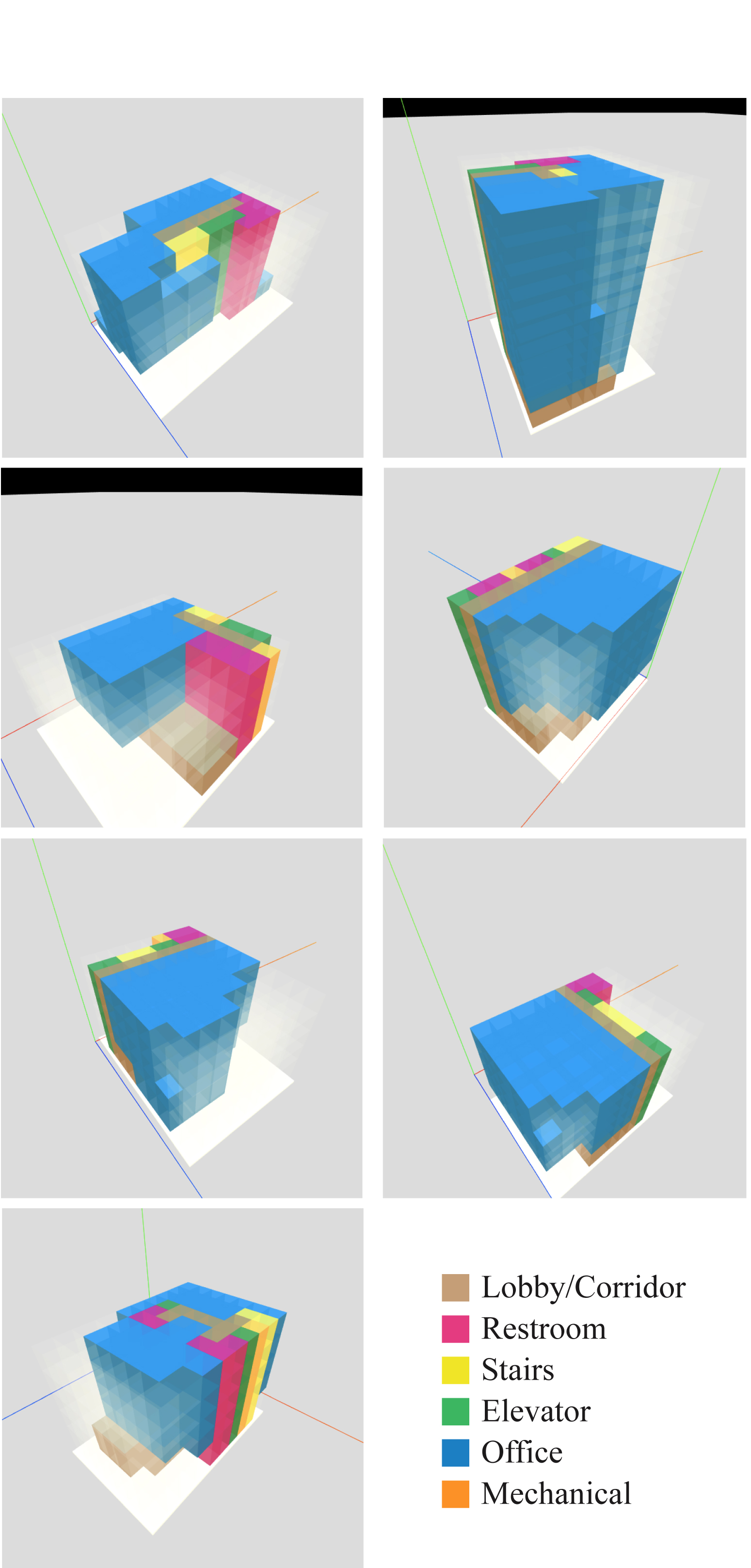}
  \caption{Continued}
  \label{fig:syn_2}
\end{figure}

\begin{figure}[h]
  \centering
  \includegraphics[width=0.5\textwidth]{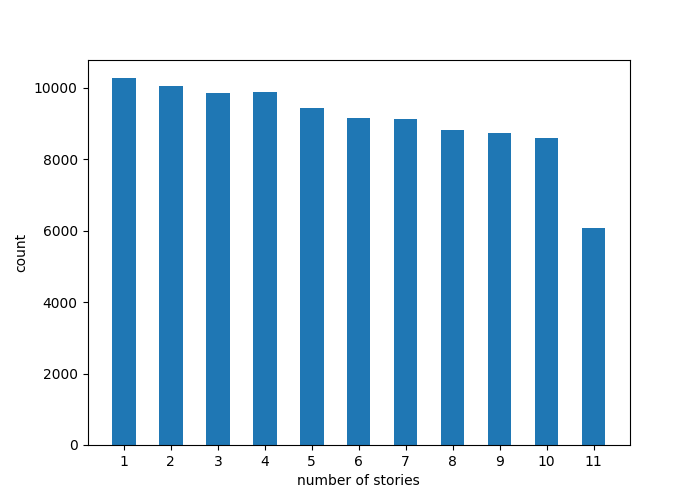}
  \caption{Number of stories distribution}
  \label{fig:floor_dist}
\end{figure}

\begin{figure}[h]
  \centering
  \includegraphics[width=0.5\textwidth]{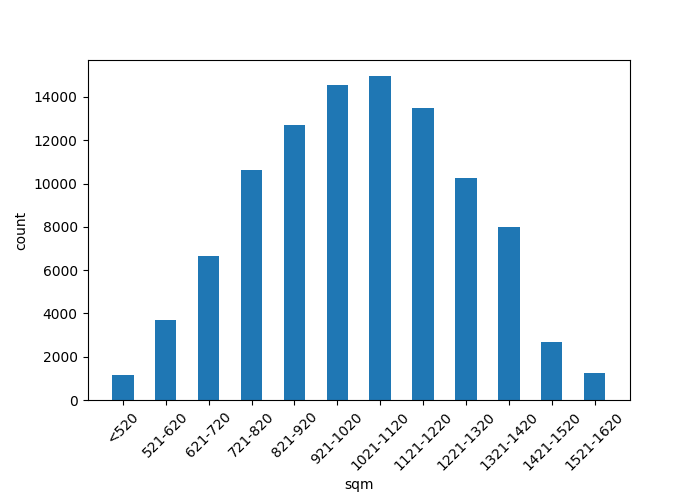}
  \caption{Site area distribution}
  \label{fig:site_dist}
\end{figure}

The synthetic dataset contains 120,000 volumetric designs that are represented in voxel graph with JSON format. They can also be converted into either conventional voxels, meshes, or solid geometries. For each design, the site conditions such as the shape and maximum height are randomly generated from the distribution shown in Figure \ref{fig:floor_dist} and \ref{fig:site_dist}. The site conditions in Building-GAN are also represented in voxel graphs as part of the input conditions. Other conditions, such as the targeted TPR and FAR, and the input program graph (bubble diagram) are all restored in JSON format.

Figure \ref{fig:syn_1} and \ref{fig:syn_2} present several examples in the synthetic dataset. We apply patterns and rules provided by professional architects to generate this dataset. For example, we can observe that some of the buildings have symmetric elevators facing each other (e.g., top left in Figure \ref{fig:syn_1}), while some corridors form left and right wings for the same set of functional spaces (e.g., bottom right in Figure \ref{fig:syn_1}. Also, the first story usually has a different layout than other stories due to the entrance and lobby requirements for commercial buildings. Moreover, the first story often has a higher ceiling compared to other floors. 

Besides the interior functionalities, building façades, i.e., the exterior shapes of buildings, is also an important aesthetic factor to consider in the architectural design process. To increase the façade's variety in this dataset, we generate story partition patterns that divide stories into groups. For instance, the top two stories in the top left in Figure \ref{fig:syn_1} have a different shape than the stories below them and therefore create a shape feature in 3D. In practice, such a shape feature can have functional purposes such as a terrace or roof garden. Please refer to Section \ref{sec:A8} Case Study for a architect-generated  building design example.

\section{Implementation Detail}

We implement our model in PyTorch. The latent dimension is 128 and the noise dimension (for both program and voxel graphs) is 32. The message passing steps in program and voxel GNN are 4 and 12 respectively. We summarize MLP specifications in Table \ref{table:MLP_spec} and the pseudo-code of our generator is given in Algorithm \ref{alg:generator}.

\begin{table*}[h]
\begin{center}
\begin{tabular}{|l|c|c|}
\hline
\multicolumn{3}{|c|}{Generator}\\
\hline
\multirow{3}{*}{Program GNN}& Encoder MLP & Dense(input=5+32, output=128)\\
 & Message MLP & Dense(input=256, output=128) \\
  & Update MLP & Dense(input=384, output=128,act=LeakyReLU) \\
\hline
\multirow{5}{*}{Voxel GNN}& Encoder MLP & Dense(input=4+32, output=128)\\
 & Message MLP & Dense(input=256+3, output=128) \\
 & Update MLP & Dense(input=256, output=128,act=LeakyReLU) \\  
 & Mask MLP & Dense(input=128, middle=64, output=2)\\
 & Attention $W_{x}$, $W_{v}$&Dense(input=128, output=128)\\ 
\hline
\hline
\multicolumn{3}{|c|}{Discriminator}\\
\hline
\multirow{6}{*}{Voxel GNN}& Feature Encoder MLP & Dense(input=4, output=128)\\
 & Label Encoder MLP & Dense(input=6, output=128) \\
 & Message MLP & Dense(input=512+3, output=256) \\
 & Update MLP & Dense(input=512, output=256,act=LeakyReLU) \\  
 & Building Decoder MLP & Dense(input=256, middle=128, output=1,act=Sigmoid)\\
 & Story Decoder MLP & Dense(input=256, middle=128, output=1,act=Sigmoid)\\
\hline
\end{tabular}
\end{center}
\caption{Model specifications.}
\label{table:MLP_spec}
\end{table*}

\begin{figure*}[h!]
  \centering
  \includegraphics[width=\textwidth]{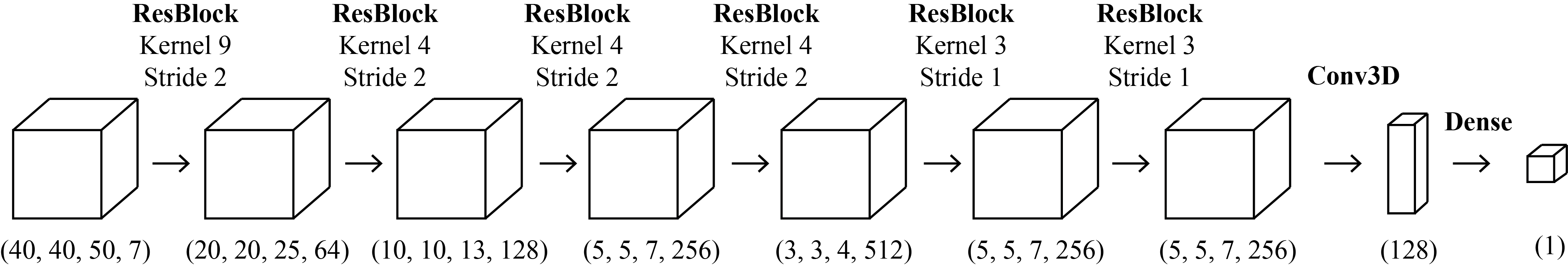}
  \caption{The network architecture of the 3D Descriptor Net used for FID computation}
  \label{fig:3ddescriptor}
\end{figure*}

\begin{figure*}[h!] 
  \centering
  \includegraphics[width=\textwidth]{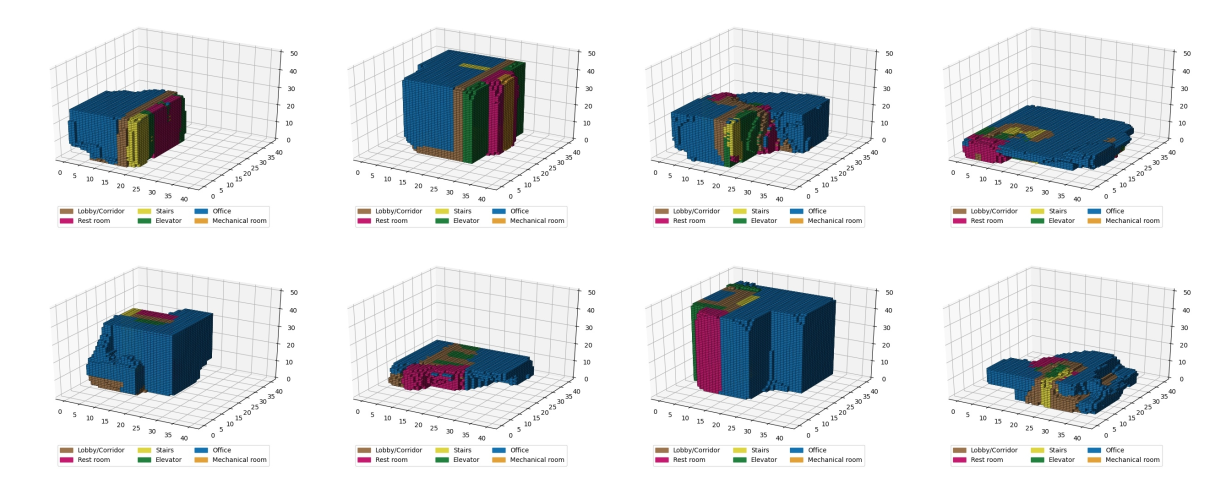}
  \caption{Synthesized examples by 3D Descriptor Net}
  \label{fig:3ddescriptorresult}
\end{figure*}

The 120,000 data is split to 96,000 data as training set and 24,000 data as test set. We train the models using two NVIDIA GV100 GPUs and an Intel i7 CPU with 16 cores. The total training epoch is 50 using Adam Optimizer ($b_{1}=0.5, \;b_{2}=0.999$) and batch size 8. The learning rates of the generator and the discriminator are both 0.0001. The generator updates its parameters every 5 discriminator update steps. 

\begin{algorithm}[h] 
  \caption{Generator}
  \label{alg:generator}
\begin{algorithmic}
    \STATE {\bfseries Input:} program graph $x$, voxel graph $v$, noise $z^{p}, z^{v}$
    \\\hrulefill
    \STATE 1: Program GNN
    \STATE $x^{0} \gets$ Program Graph Encoder($x, z^{p}$)
    \FOR{$t=0$ {\bfseries to} $T-1$}
    \STATE $x^{t+1} \gets$ Program Graph Update($x^{t}$)
    \ENDFOR
    \\\hrulefill
    \STATE 2: Voxel GNN
    \STATE $v^{0} \gets$ Voxel Graph Encoder($v, z^{v}$) + PE($v$)
    \STATE \_ , \_ , $v^{0} \gets$ Pointer($v^{0}, x^{T}$)
    \FOR{$t=0$ {\bfseries to} $T-1$}
    \STATE $v^{t+1} \gets$ Voxel Graph Update($v^{t}$)
    \IF{$t<T-1$}
    \STATE \_ , \_ , $v^{t+1} \gets$ Pointer($v^{t}, x^{T}$)  \dots (Optional)
    \ENDIF
    \ENDFOR
    \STATE $mask^{T}, att^{T}, v^{T} \gets$ Pointer($v^{T}, x^{T}$)
\end{algorithmic}
\end{algorithm}

\section{3D Descriptor Network for FID score}

The backbone of the inference network used for FID computation is 3D Descriptor Net. We replace all convolution layers with 6 residual blocks due to the higher complexity of our data and insert a dense layer to convert the 128 dimension embedding to a scalar energy value. The model architecture is illustrated in Figure \ref{fig:3ddescriptor}. We train the model for 400 epochs when the average mean square error decreases to 0.0121. Some synthesized volumetric designs generated unconditionally by the trained 3D Descriptor Network are visualized in Figure \ref{fig:3ddescriptorresult}.

\begin{figure*}[h]
  \centering
  \includegraphics[width=0.85\textwidth]{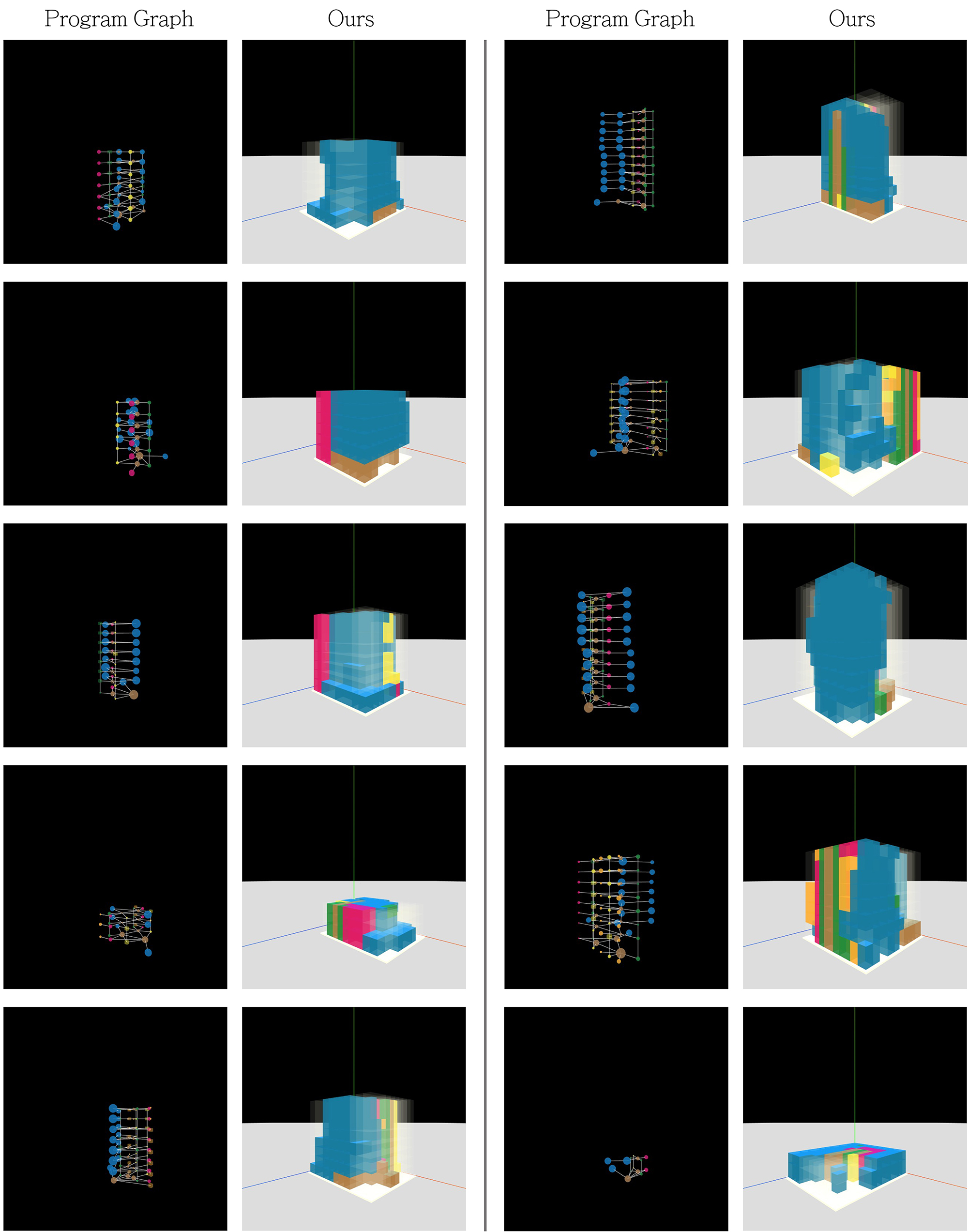}  
  \caption{More results generated from Building-GAN. Left: Input Program Graph. Right: Output Volumetric Design.}
  \label{fig:more1}
\end{figure*}

\section{Additional Results}

\subsection{Program Graph to Volumetric Design}

Figure \ref{fig:more1} to \ref{fig:more3} shows additional results generated by Building-GAN given the program graphs (bubble diagrams).

\begin{figure*}[h]
  \centering
  \includegraphics[width=0.95\textwidth]{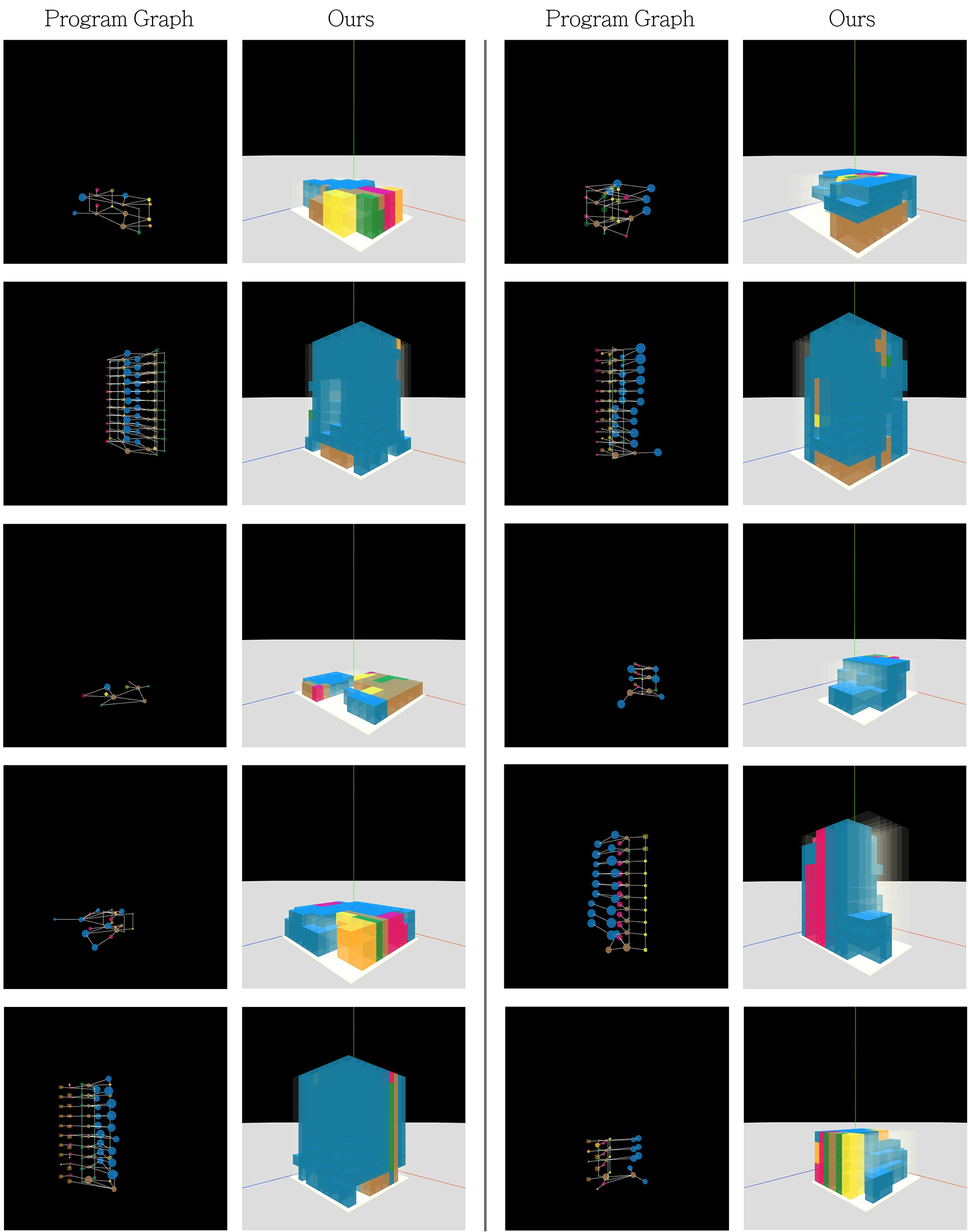}  
  \caption{Continued.}
  \label{fig:more2}
\end{figure*}

\begin{figure*}[h]
  \centering
  \includegraphics[width=0.95\textwidth]{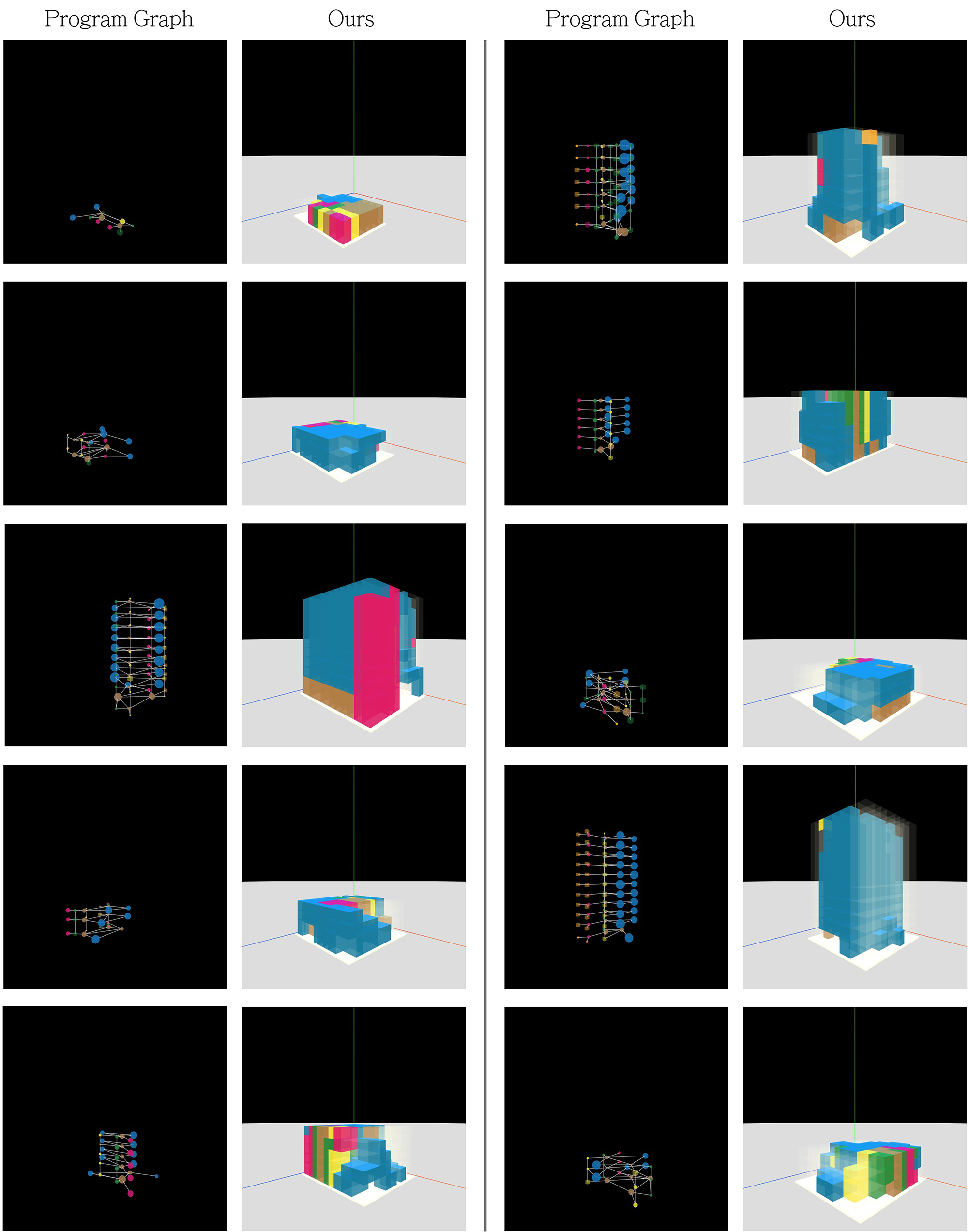}  
  \caption{Continued.}
  \label{fig:more3}
\end{figure*}

\subsection{Floor Plans from Volumetric Design}
Figure \ref{fig:layou1} to \ref{fig:layou3} provide examples of floor plan layouts sliced from volumetric designs generated by Building-GAN.

\begin{figure*}[h]
  \centering
  \includegraphics[width=0.95\textwidth]{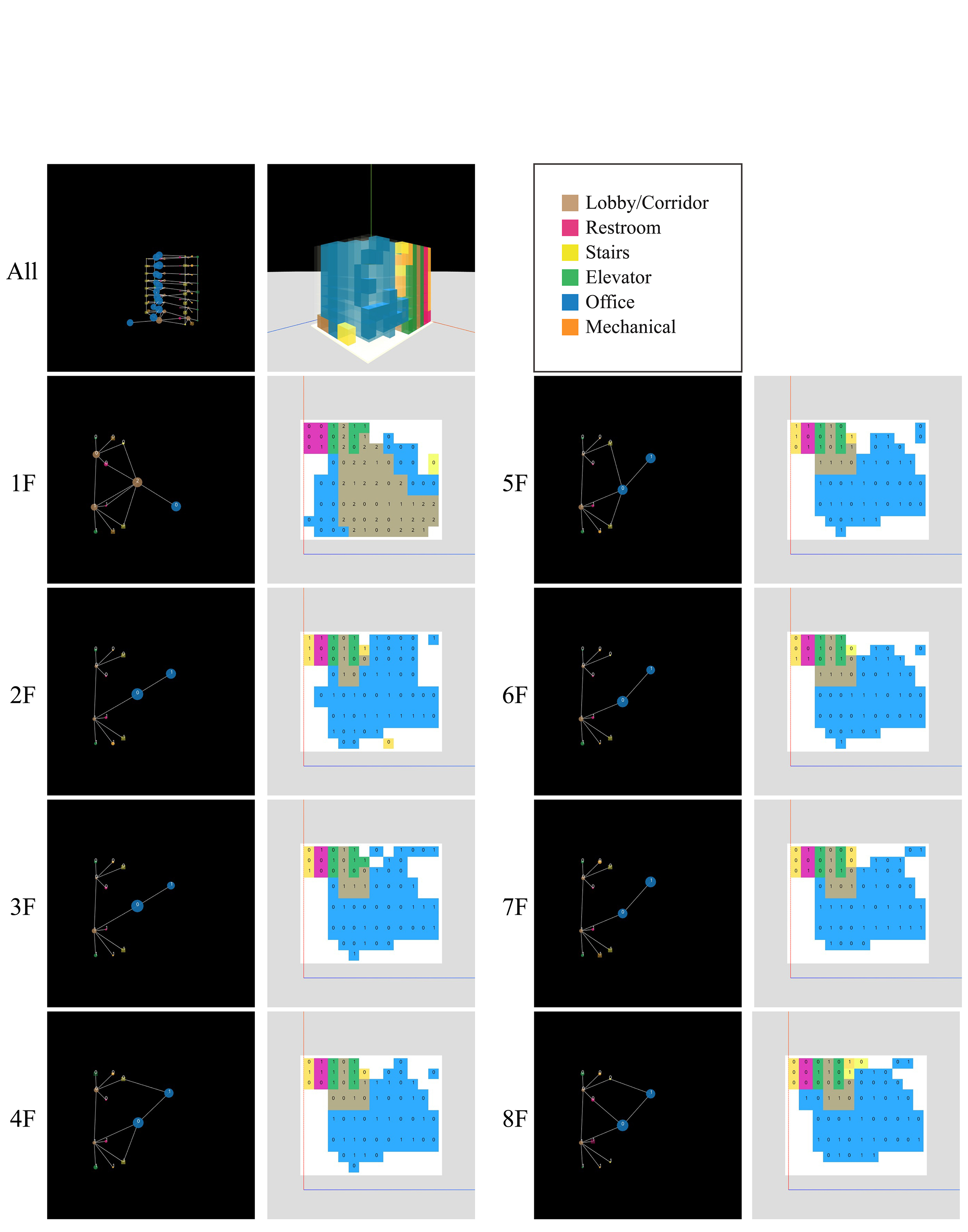}  
  \caption{Floor plan layouts from volumetric design.}
  \label{fig:layou1}
\end{figure*}

\begin{figure*}[h]
  \centering
  \includegraphics[width=0.95\textwidth]{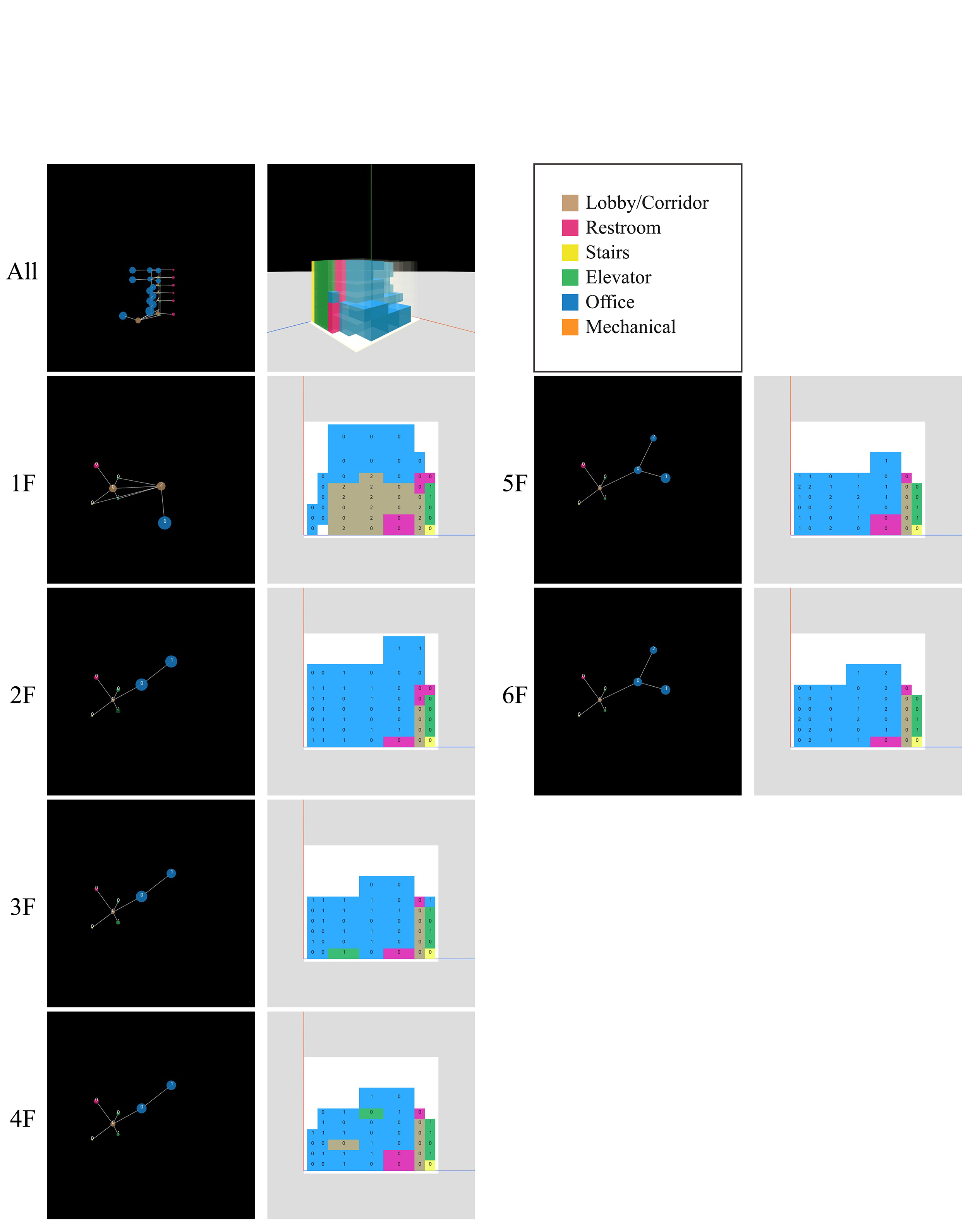}  
  \caption{Continued.}
  \label{fig:layou2}
\end{figure*}

\begin{figure*}[h]
  \centering
  \includegraphics[width=0.95\textwidth]{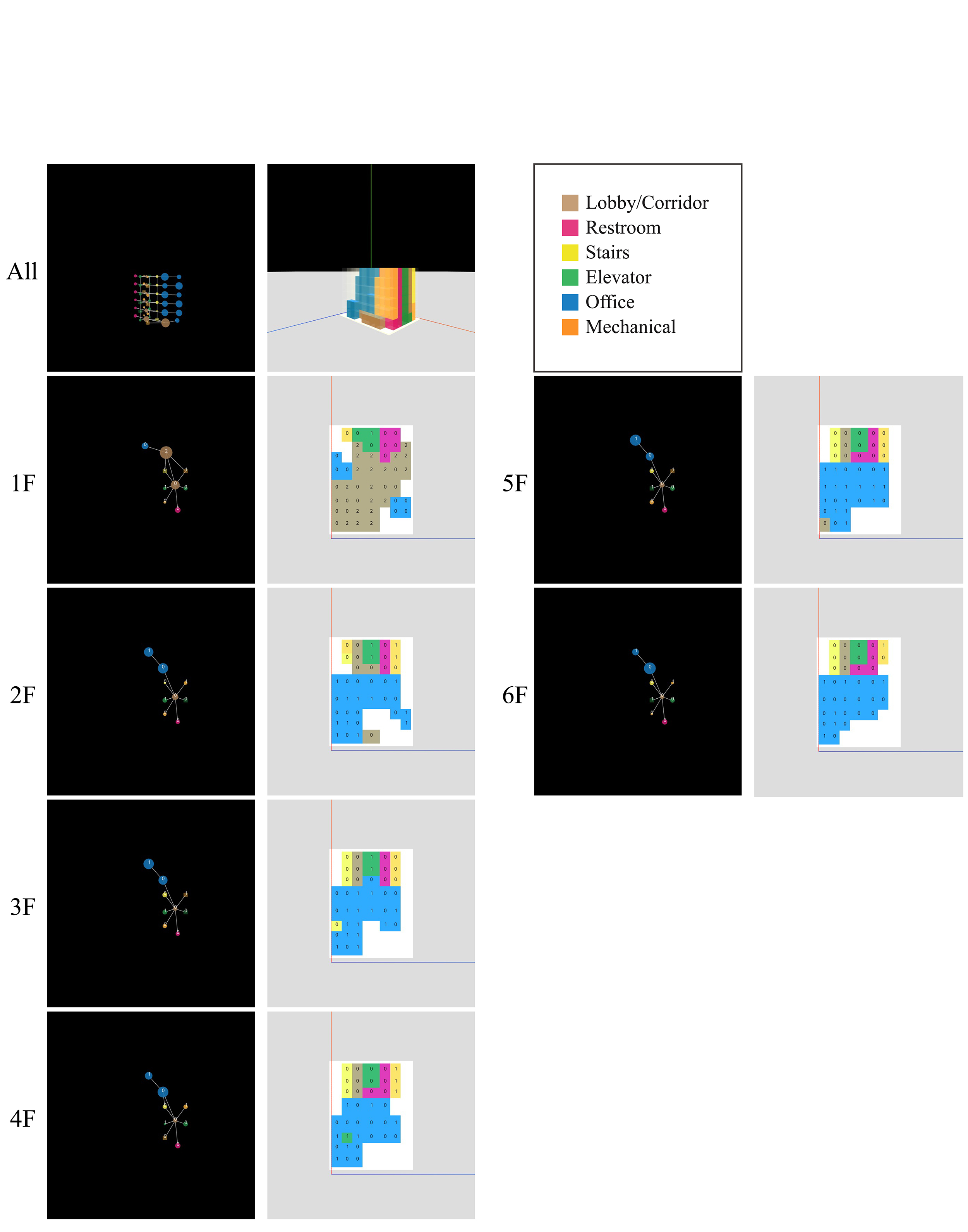}  
  \caption{Continued.}
  \label{fig:layou3}
\end{figure*}


\section{FAR and TPR Study}

We extend our comparative study and ablation study conducted in Section 5 on two more metrics: Floor Area Ratio (FAR) distance and Target Program Ratio (TPR) accuracy. The results are shown in Table~\ref{table:baseline_comparison-far} and Table~\ref{table:ablation-far} respectively. We use the difference between the actual FAR and the target FAR over the target FAR to calculate the FAR distance for each design. The TPR accuracy of each design is calculated as 1 minus the sum of the absolute difference between the actual program ratio and the target program ratio of each room type. All results are averaged over 10,000 samples. 

Table \ref{table:baseline_comparison-far} shows that TPR improves slightly when we increase the number of message passing layers, while FAR slightly decreases. We can also see that both FAR and TPR improve while raising the frequency for the pointer module. Table \ref{table:ablation-far} shows that building discriminator helps significantly on both FAR and TPR since these two values are building-level properties.

\begin{table}[h]
\begin{center}
\begin{tabular}{|l|c|c|c|}
\hline
Method & Parameter & FAR & TPR\\
\hline\hline
House-GAN & - & 0.853 & 0.528\\
\hline
Ours & - & \cellcolor{lightgray} 1.210  & \cellcolor{lightgray} 0.772\\
\hline \hline
\multirow{5}{*}{
\begin{tabular}{@{}l@{}} Voxel Layer \\ (Pointer Frequency \\ = every 2 steps)\end{tabular}
} & 4 & 1.075 & 0.744\\
& 6 & 1.159 & 0.759\\
& 8 & 1.117 & 0.749\\
& 10 & 1.144 & 0.754\\
& 12 & \cellcolor{lightgray} 1.210  & \cellcolor{lightgray} 0.772\\ \hline  
\multirow{4}{*}{
\begin{tabular}{@{}l@{}}Pointer Frequency \\ (Voxel Layer = 12)\end{tabular}
} & first + last & 2.026 & 0.512\\ 
& every 6 steps& 1.290 & 0.743\\ 
& every 3 steps & 1.393 & 0.734\\
& every 2 steps & \cellcolor{lightgray} 1.210 & \cellcolor{lightgray} 0.772\\  
\hline
\end{tabular}
\end{center}
\caption{Quantitative evaluation using FAR distance and TPR accuracy. We compare our baseline model to House-GAN. We also experiment baseline models with different numbers of voxel layer and pointer frequencies.}
\label{table:baseline_comparison-far}
\end{table}

\begin{table}[h]
\begin{center}
\begin{tabular}{|l|c|c|}
\hline
Ablation Study & FAR & TPR\\
\hline\hline
Ours & 1.210 & 0.772\\ 
\hline\hline
Story discriminator only & 2.026 & 0.207 \\
Building discriminator only & 1.247 & 0.717\\
\hline\hline
No PE & 1.260 & 0.731\\ 
No PE + No RP &  1.065 & 0.673\\ 
\hline
\end{tabular}
\end{center}
\caption{Ablation study results of FAR distance and TPR accuracy on discriminator, positional encoding (PE), and relative position (RP).}
\label{table:ablation-far}
\end{table}

In our best model, TPR is around 77\% accurate, and FAR is around 1.2, which means about 15-20\% error on each floor. Both numbers are acceptable given there are no explicit loss terms applied to these two conditions. However, we found that these two values are relatively trivial to learn compared to connectivity. First, FAR is highly dependent on the number of stories, i.e., higher buildings have higher FAR. Since the number of stories is given by the program graph, we can expect the FAR of the generated designs to have similar distribution if the model learns the overall shape distribution. Secondly, although TPR values vary in each program graph, they are following some prior design rules. For example, there should not be any building with elevators covering more than 30\% of the floor area. Therefore, as long as the model learns the general type distribution on voxel graphs, TPR will not be significantly off. Based on the analysis above, we choose not to use TPR and FAR as our evaluation metrics.
\section{User Study Details}

\begin{figure}[h]
  \centering
  \includegraphics[width=0.4\textwidth]{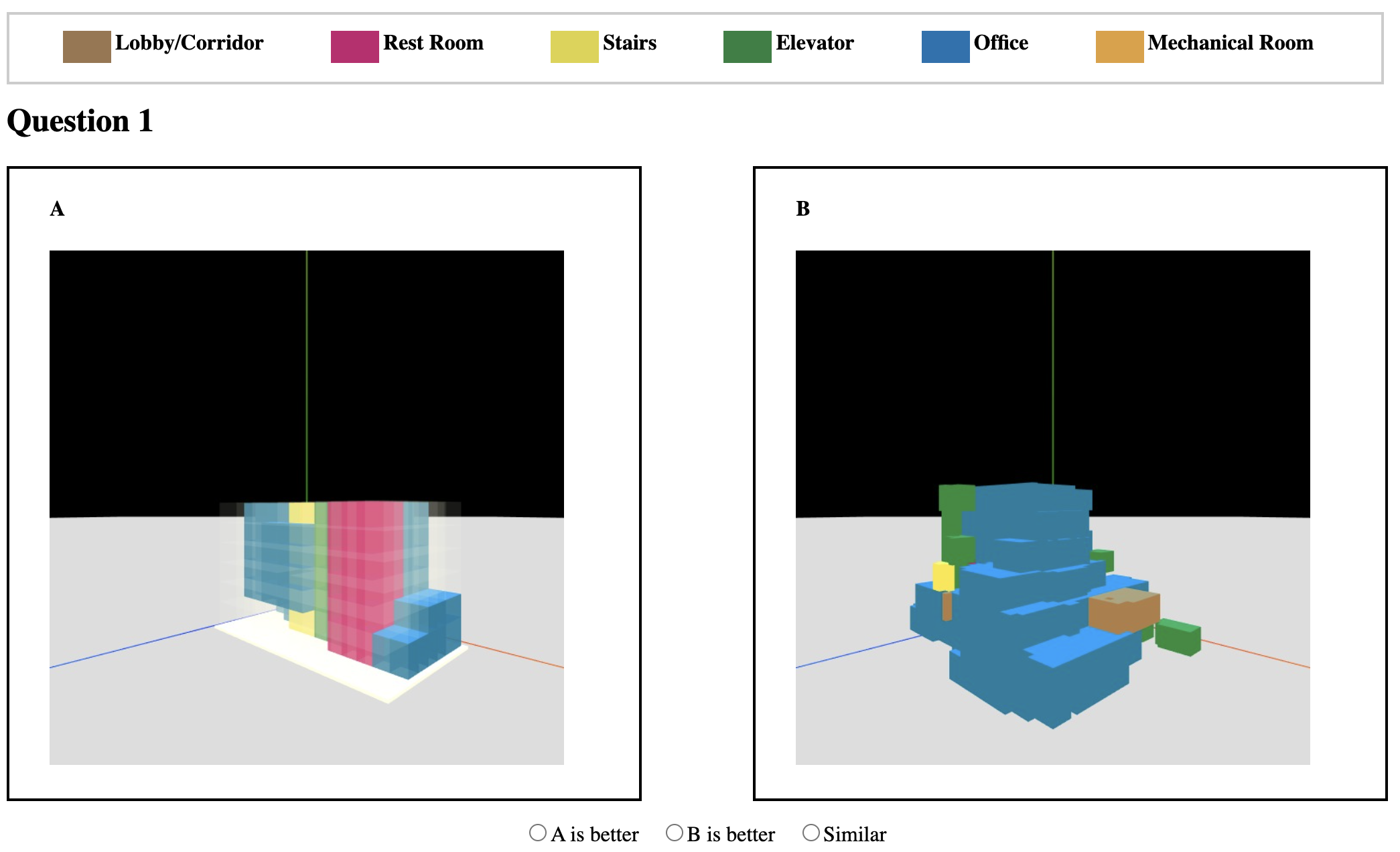}
  \caption{A screenshot of our user study interface: the annotations appear on the top, followed by a pair of generated samples for each question.}
  \label{fig:user_study_1}
\end{figure}

\begin{figure}[h]
  \centering
  \includegraphics[width=0.4\textwidth]{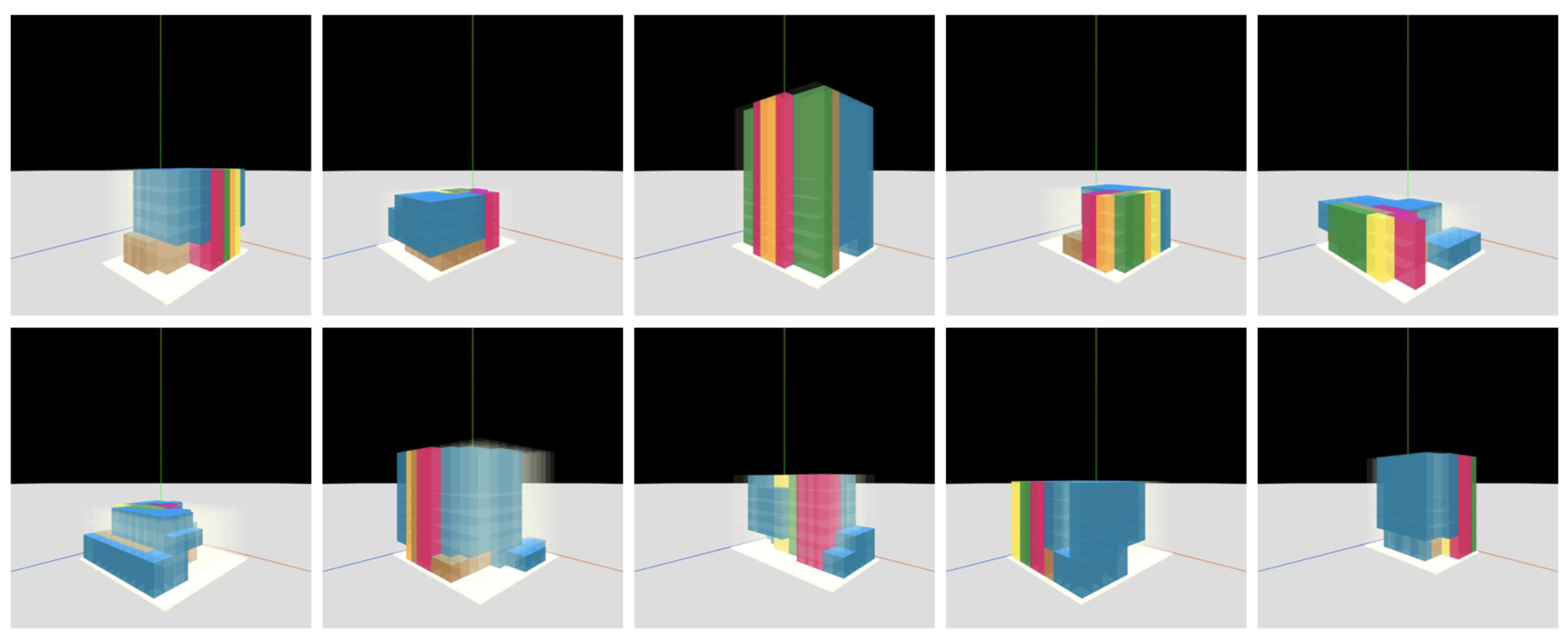}
  \caption{A screenshot of the reference page of our user study.}
  \label{fig:user_study_2}
\end{figure}

Figure~\ref{fig:user_study_1} shows a screenshot of our user study interface. A
subject is presented with annotations of the room types on the top, followed by a pair of generated volumetric designs for each question. In addition, a reference page is provided to show a set of ground-truth volumetric designs, as shown in Figure~\ref{fig:user_study_2}. Each subject is given 48 questions and asked to choose one of the three possible answers (“A is better”, “B is better”, “Similar”), where the entire session takes around 20 minutes to be completed. The generated sample pairs come from pairs of randomly selected models among HouseGAN, Building-GAN, and ground-truth. We enforce that each possible pair of models is selected exactly 16 times during the entire session.

\section{Case Study}
\label{sec:A8}
\subsection{Workflow and Setup}
We invite an architect to go through the design process using Building-GAN and see if he can speed up the current workflow and create good volumetric designs. The typical workflow using conventional tools is as follow:
\begin{enumerate}
    \small
    \item Collect the project information, including site boundary, height restriction, FAR, and the desired program requirements (TPR) from the client.
    \item Create the bubble diagram and drawing the partitions or grids on the site.
    \item Make 2D drawings, 3D models, or physical models using foam or foam board to create volumetric designs.
    \item Arrange functional space, also called “core” in the architectural industry, including elevators, stairs, restrooms, mechanical rooms.
    \item Repeat from step 2 to 4 until the design is satisfying.
    \item Draw and model the interior details as well as the façade design.
    \item Create drawings, renderings, and presentation slides.
    \item Present to the client and repeating from any of the previous steps if the client disagrees with the current proposal.
\end{enumerate}

Based on the evaluation from a professional architecture firm, this process usually takes about two to three weeks with three architectural designers. Building-GAN is designed for speeding up the loop from steps 2 to 5, where the user can quickly specify the requirements and explore the design options interactively by modifying the bubble diagram and voxel partitions. In this case study, we ask the architect to create the volumetric design using Building-GAN and complete step 6 and 7. We measure the time he spends on each step and collect feedback from him.

\subsection{Results}

\begin{table}[h]
\begin{center}
\begin{tabular}{|l|c|c|}
\hline

Tasks & Time (hrs)\\
\hline 
Draw program graph and site partitions & \cellcolor{lightgray} 0.2\\
Explore the design options & \cellcolor{lightgray} 0.25\\
Modify the generated volumetric design & 1\\
Design the interior and facade & 8\\
Choose materials and set up for rendering & 2.5\\
Render images from different angles & 3\\
\hline
\end{tabular}
\end{center}
\caption{Time spent by the architect in each task of the case study.}
\label{table:time_task}
\end{table}

As shown in Table \ref{table:time_task}, the architect uses only 10 minutes to set up the problem and 15 minutes to find the desired volumetric design. The total labor hours are reduced to less than two days. Although more adjustment and documentation work might be needed for a formal presentation to the client, the architect does feel a significant speed up on exploring valid design options while still being able to drive the design idea by himself. This is an important feedback for us--Building-GAN can help complete professional level tasks while making the architect feel their creativity is not limited nor replaced by providing real-time interaction with the system.

\begin{figure}[h]
  \centering
  \includegraphics[width=0.35\textwidth]{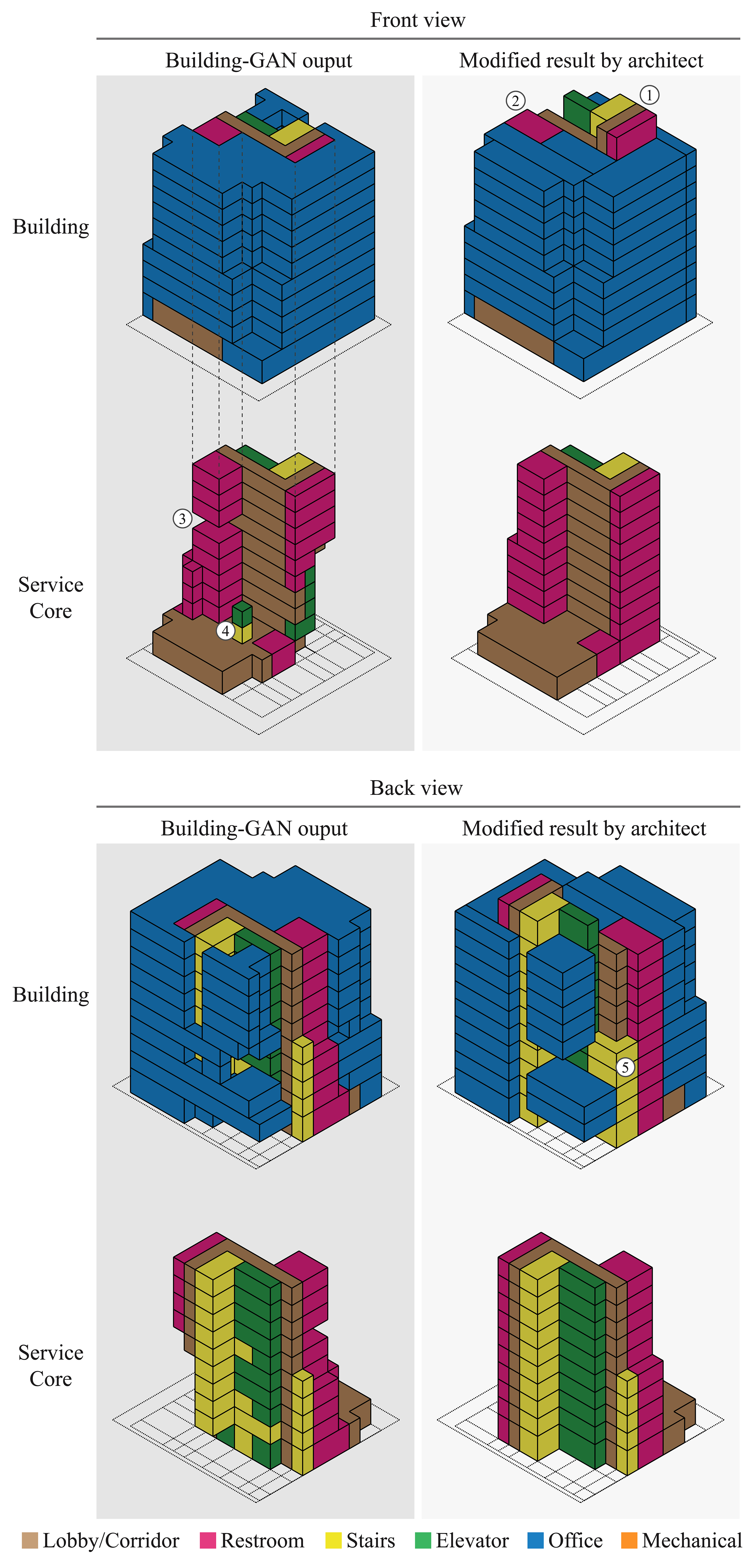}  
  \caption{Comparison between the Building-GAN output and the modified volumetric design by the architect. There are five major types of modifications: 1) extending the stairs and elevators to the roof; 2) adjusting the boundaries of rooms for alignments; 3) filling the gaps; 4) removing redundant rooms; 5) expanding or shrinking the required space.}
  \label{fig:modification}
\end{figure}

In addition, we found that the architect can easily modify the flawed results generated from Building-GAN. As long as the generated design is conceptually or roughly correct, meaningful, or inspiring, the architect can fix those flaws and move on to the next stage. Examples can be found in Figure \ref{fig:modification}. The rendered images of the completed building design are shown in Figure \ref{fig:case1} and \ref{fig:case2}.

\section{Failed Attempts}

\begin{figure}[h]
  \centering
  \includegraphics[width=.45\textwidth]{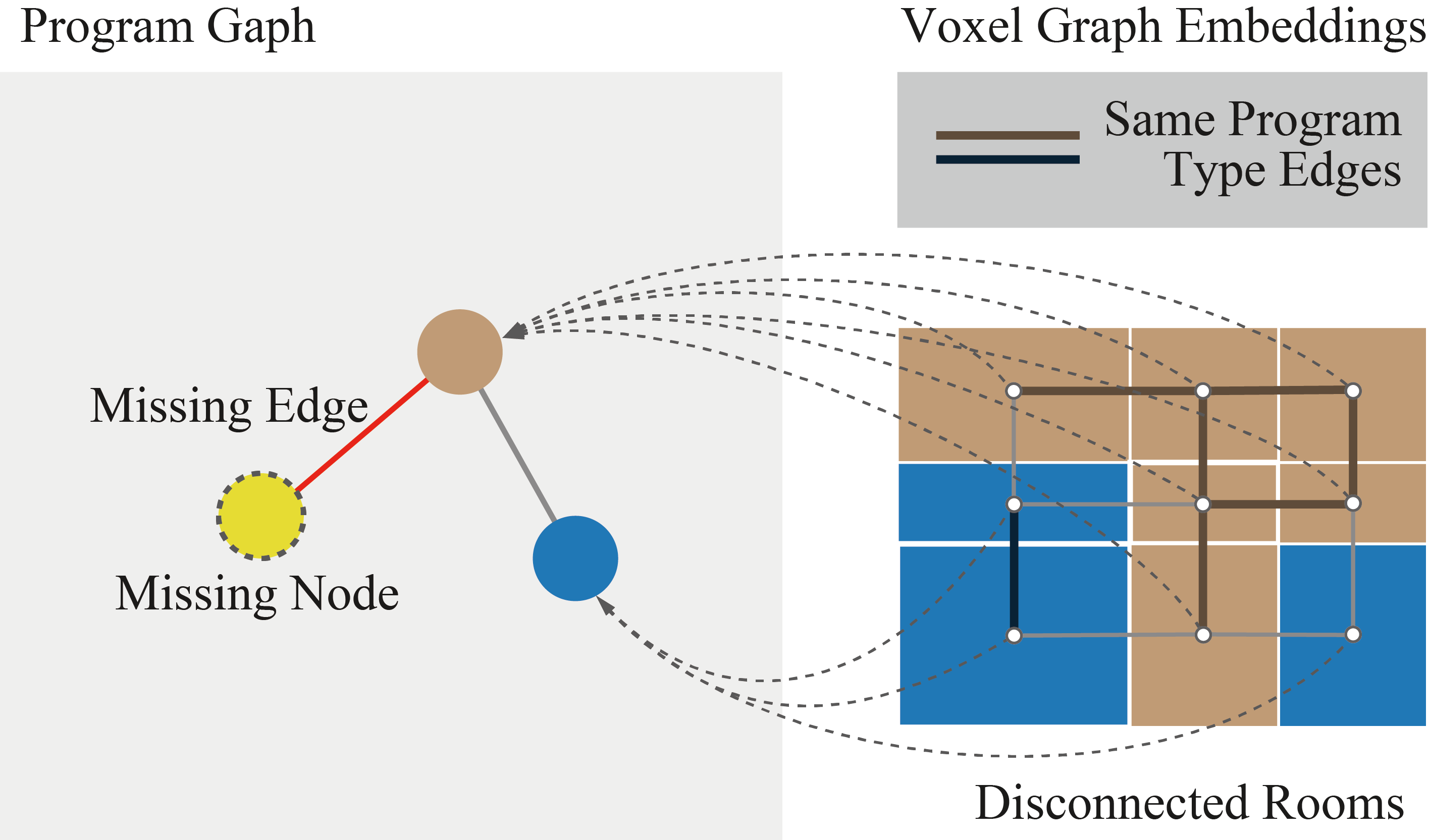}  
  \caption{We tried aggregating the voxel embeddings back to program graph, but the GAN training is not stable.}
  \label{fig:failureback}
\end{figure}

The failure cases shown in the paper is very likely because our discriminator only observes the program types on the voxel nodes to evaluate whether the entire design is realistic or not. Since the program graph is not taken as input, the discriminator cannot critic on missing nodes and missing edges on the program graph, nor the fragmented layouts (disconnected rooms). To resolve this issue, we experiment with aggregating the voxel node embeddings back to the program nodes. More specifically, the attention output from the generator indicates the program node selected by each voxel node, and we aggregate the voxel node embeddings that point to the same program node together as illustrated in Figure~\ref{fig:failureback}. The aggregated embedding is expected to provide information that tells if there's no voxel node pointing to the program node and if the voxel nodes form connected rooms. Though the attempt fails to generate promising results, we think it is helpful to share the lessons learned. 

We experiment with two approaches that apply this concept. First, a program discriminator can be designed by aggregating the voxel node embeddings to the program nodes. Then the program discriminator classifies each program node. Additional message passing can be added before the aggregation based on "same program type edges (shown in \ref{fig:failureback})" or after the aggregation to identify missing edges. However, the GAN training becomes unstable after adding the program discriminator. It is possibly due to the dynamic aggregation from the generated designs. This operation is also non-differentiable with respect to the attention output computed by the generator since it is used as aggregation indices. The other approach is to take the voxel embeddings from the generator and apply an explicit link prediction loss using contrastive learning. The positive and negative pairs are sampled from program edges and missing edges respectively.  Cosine similarity and InfoNCE loss are used to encourage connected program nodes to have more similar aggregated embeddings. The link prediction loss is added to the GAN loss with a specific weight. Our experiments also show unstable GAN training after we add the link prediction loss. From the failed attempts, we learn that it is hard to provide stable gradients by aggregating the voxel embeddings back to program nodes. We leave this challenge for future work. 

\begin{figure*}[h]
  \centering
  \includegraphics[width=0.7\textwidth]{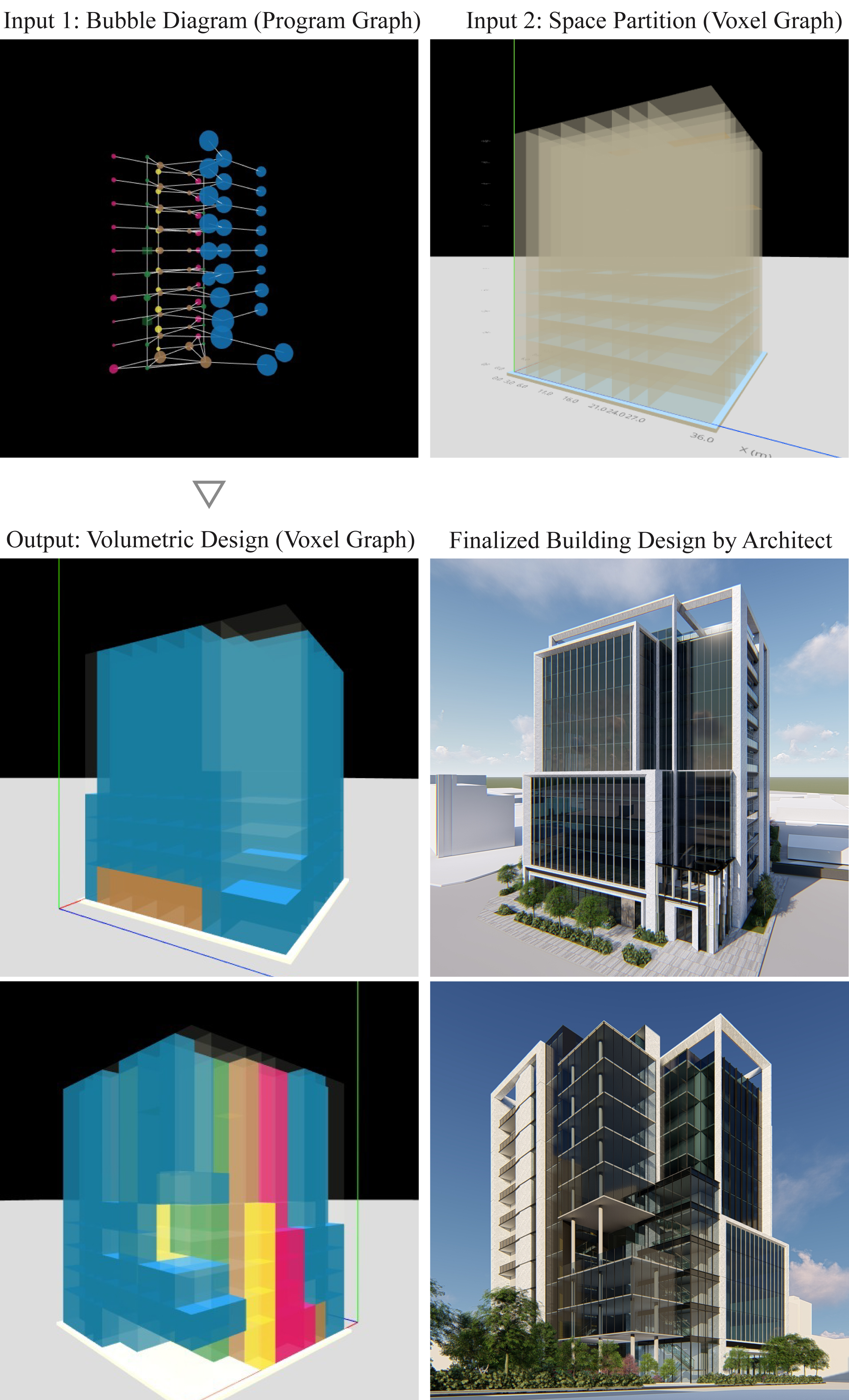}  
  \caption{Results of the design process by the architect using Building-GAN.}
  \label{fig:case1}
\end{figure*}

\begin{figure*}[h]
  \centering
  \includegraphics[width=0.7\textwidth]{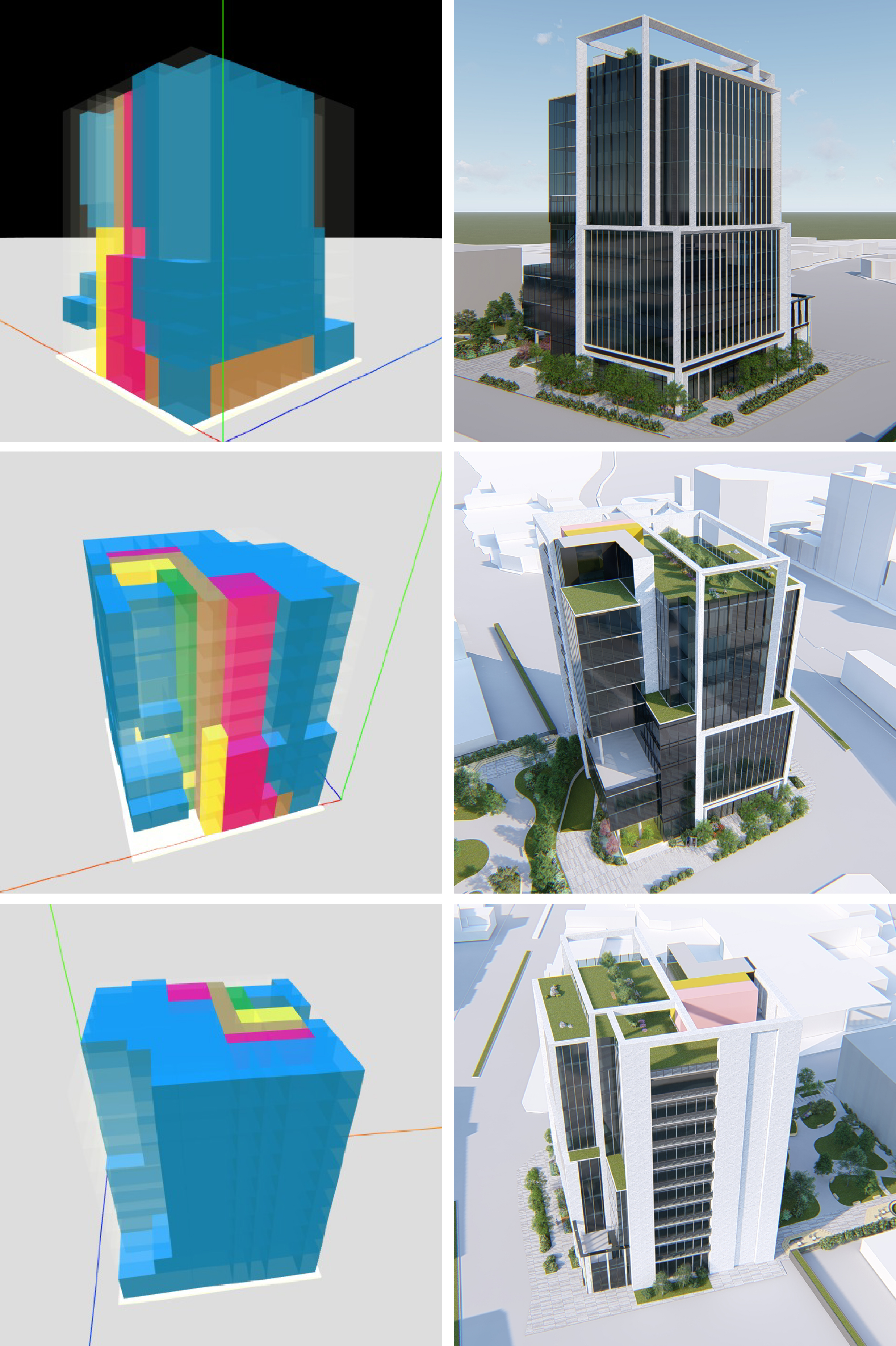}  
  \caption{Continued.}
  \label{fig:case2}
\end{figure*}


\end{document}